\documentclass{article}
\usepackage{pdfpages}

\usepackage{times}
\usepackage{multicol}
\usepackage{graphicx} 
\usepackage{subfigure} 
\usepackage{mathrsfs}
\usepackage{natbib}

\usepackage{algorithm}
\usepackage{algorithmic}
\usepackage{hyperref}
\usepackage{booktabs}
\usepackage{footmisc}

\usepackage[title]{appendix}

\usepackage[accepted]{icml2017} 

\usepackage{macros}
\usepackage{todonotes}

\usepackage{changepage}
\newlength{\offsetleft}
\newlength{\offsetright}
\newenvironment{widepage}{\begin{adjustwidth}{-\offsetleft}{-\offsetright}%
    \addtolength{\textwidth}{\offsetleft+\offsetright}}%
{\end{adjustwidth}}

\icmltitlerunning{Real-Time Adaptive Image Compression}

\begin{document} 

\twocolumn[
\icmltitle{Real-Time Adaptive Image Compression}



\icmlsetsymbol{equal}{*}
\begin{icmlauthorlist}
\icmlauthor{Oren Rippel}{equal,wo}
\icmlauthor{Lubomir Bourdev}{equal,wo}
\end{icmlauthorlist}

\icmlaffiliation{wo}{WaveOne Inc., Mountain View, CA, USA}

\icmlcorrespondingauthor{Oren Rippel}{oren@wave.one}
\icmlcorrespondingauthor{Lubomir Bourdev}{lubomir@wave.one}

\icmlkeywords{boring formatting information, machine learning, ICML}

\vskip 0.2in
]



\printAffiliationsAndNotice{\icmlEqualContribution} 


\begin{abstract}
We present a machine learning-based approach to lossy image compression which outperforms all existing codecs, while running in real-time.

Our algorithm typically produces files 2.5 times smaller than JPEG and JPEG 2000, 2 times smaller than WebP, and 1.7 times smaller than BPG on datasets of generic images across all quality levels. At the same time, our codec is designed to be lightweight and deployable: for example, it can encode or decode the Kodak dataset in around 10ms per image on GPU. 

Our architecture is an autoencoder featuring pyramidal analysis, an adaptive coding module, and regularization of the expected codelength. We also supplement our approach with adversarial training specialized towards use in a compression setting: this enables us to produce visually pleasing reconstructions for very low bitrates.
\end{abstract}


\section{Introduction}
\label{sec:introduction}
Streaming of digital media makes 70\% of internet traffic, and is projected to reach 80\% by 2020~\cite{CISCOForecast}. However, it has been challenging for existing commercial compression algorithms to adapt to the growing demand and the changing landscape of requirements and applications. While digital media are transmitted in a wide variety of settings, the available codecs are ``one-size-fits-all'': they are hard-coded, and cannot be customized to particular use cases beyond high-level hyperparameter tuning.

In the last few years, deep learning has revolutionized many tasks such as machine translation, speech recognition, face recognition, and photo-realistic image generation. Even though the world of compression seems a natural domain for machine learning approaches, it has not yet benefited from these advancements, for two main reasons. First, our deep learning primitives, in their raw forms, are not well-suited to construct representations sufficiently compact. Recently, there have been a number of important efforts by \citet{toderici2015variable,toderici2016full}, \citet{twitter2016}, \citet{nyu2016}, and \citet{johnston2017improved} towards alleviating this: see Section \ref{sec:related}. Second, it is difficult to develop a deep learning compression approach sufficiently efficient for deployment in environments constrained by computation power, memory footprint and battery life. 

\begin{table}[b]
\footnotesize
\begin{tabular}[t]{lrrrr}
    \toprule
    {\bf Codec}          & {\bf\shortstack{RGB file\\ size (kb)}} & {\bf\shortstack{YCbCr file\\ size (kb)}} & {\bf \shortstack{Encode\\ time (ms)}} & {\bf \shortstack{Decode\\ time (ms)}}\\
    \toprule
    Ours               &     21.4 (100\%) &  17.4 (100\%) &     8.6$^*$ &         9.9$^*$ \\
    \cmidrule(r){1-5}
    JPEG               &      65.3 (304\%) &  43.6 (250\%) &     18.6 &      13.0 \\
    JP2                &      54.4 (254\%) &  43.8 (252\%) &    367.4 &      80.4 \\
    WebP               &      49.7 (232\%) &  37.6 (216\%) &     67.0 &      83.7 \\
    \newline
\end{tabular}
\normalsize
\caption{Performance of different codecs on the RAISE-1k $512\times 768$ dataset for a representative MS-SSIM value of 0.98 in both RGB and YCbCr color spaces. Comprehensive results can be found in Section \ref{sec:results}. $^*$We emphasize our codec was run on GPU.}
\label{tab:results}
\end{table}

\begin{figure*}[t!]
\centering
\begin{multicols}{4}
\centering
\includegraphics[width=0.25\textwidth,trim=0cm 0cm 0cm 0cm,clip]{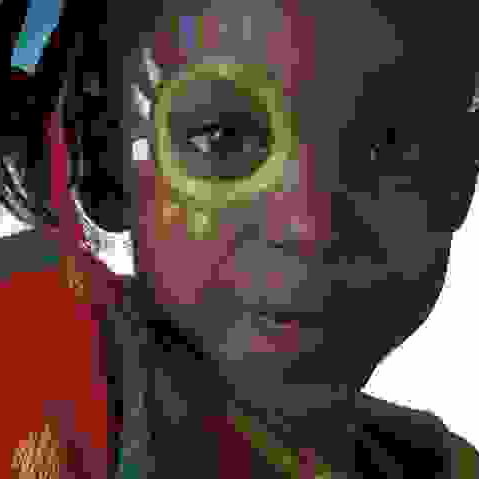}\\
{\bf JPEG}\\ 0.0826 BPP (7.5\% bigger)
\includegraphics[width=0.25\textwidth,trim=0cm 0cm 0cm 0cm,clip]{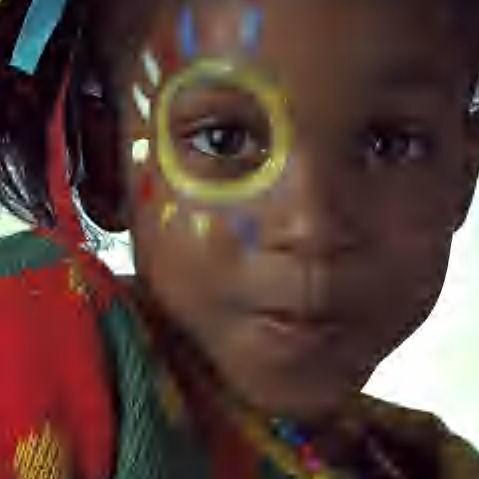}\\
{\bf JPEG 2000}\\ 0.0778 BPP
\includegraphics[width=0.25\textwidth,trim=0cm 0cm 0cm 0cm,clip]{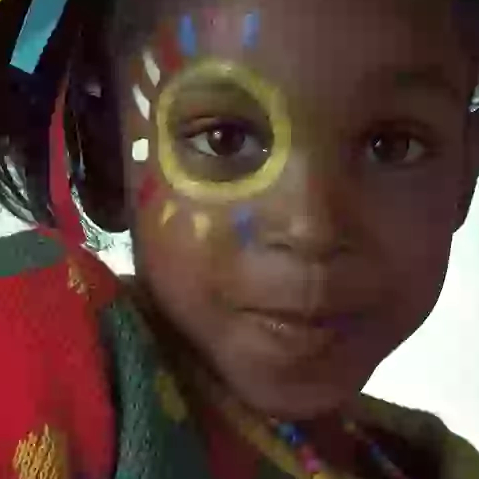}\\
{\bf WebP}\\ 0.0945 BPP (23\% bigger)
\includegraphics[width=0.25\textwidth,trim=0cm 0cm 0cm 0cm,clip]{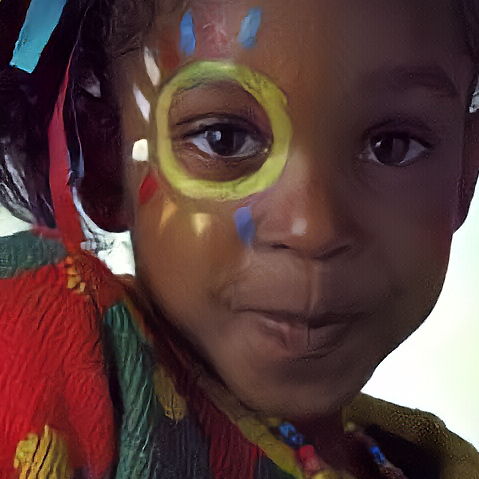}\\
{\bf Ours}\\ 0.0768 BPP
\end{multicols}
\vspace{-0.3in}
\begin{multicols}{4}
\centering
\includegraphics[width=0.25\textwidth,trim=0cm 0cm 0cm 0cm,clip]{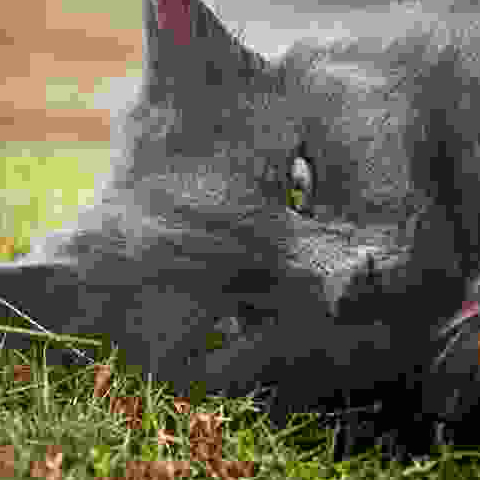}\\
{\bf JPEG}\\ 0.111 BPP (10\% bigger)
\includegraphics[width=0.25\textwidth,trim=0cm 0cm 0cm 0cm,clip]{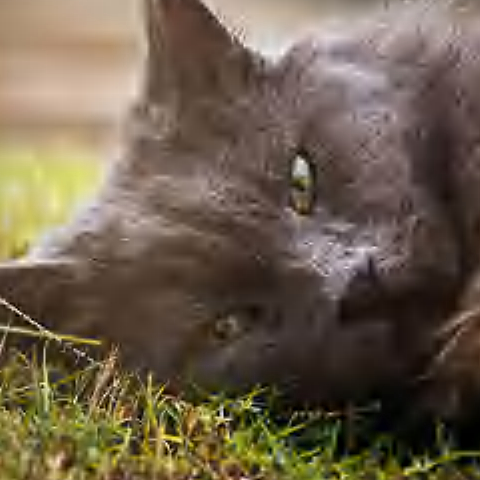}\\
{\bf JPEG 2000}\\ 0.102 BPP
\includegraphics[width=0.25\textwidth,trim=0cm 0cm 0cm 0cm,clip]{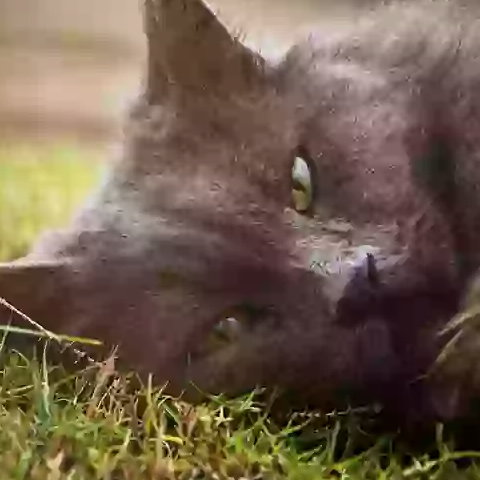}\\
{\bf WebP}\\ 0.168 BPP (66\% bigger)
\includegraphics[width=0.25\textwidth,trim=0cm 0cm 0cm 0cm,clip]{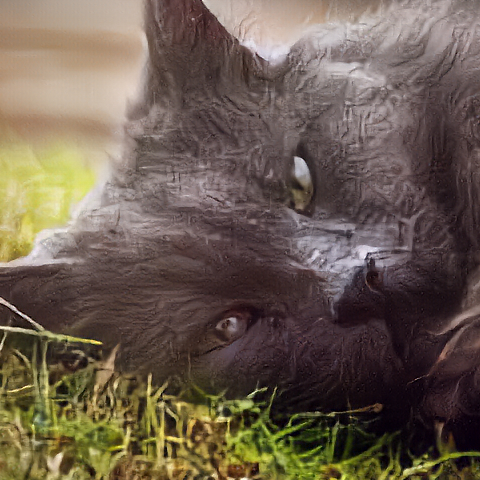}\\
{\bf Ours}\\ 0.101 BPP
\end{multicols}
\vspace{-0.2in}
\caption{Examples of reconstructions by different codecs for very low bits per pixel (BPP) values. The uncompressed size is 24 BPP, so the examples represent compression by around 250 times. We reduce the bitrates of other codecs by their header lengths for fair comparison. For each codec, we search over bitrates and present the reconstruction for the smallest BPP above ours. WebP and JPEG were not able to produce reconstructions for such low BPP: the reconstructions presented are for the smallest bitrate they offer. More examples can be found in the appendix.}
\label{fig:examples}
\vspace{-0.1in}
\end{figure*}

In this work, we present progress on both performance and computational feasibility of ML-based image compression. 

Our algorithm outperforms all existing image compression approaches, both traditional and ML-based: it typically produces files 2.5 times smaller than JPEG and JPEG~2000 (JP2), 2 times smaller than WebP, and 1.7 times smaller than BPG on the Kodak PhotoCD and RAISE-1k $512\times 768$ datasets across all of quality levels. At the same time, we designed our approach to be lightweight and efficiently deployable. On a GTX 980 Ti GPU, it takes around 9ms to encode and 10ms to decode an image from these datasets: for JPEG, encode/decode times are 18ms/12ms, for JP2 350ms/80ms and for WebP 70ms/80ms. Results for a representative quality level are presented in Table \ref{tab:results}.

To our knowledge, this is the first ML-based approach to surpass all commercial image compression techniques, and moreover run in real-time.

We additionally supplement our algorithm with adversarial training specialized towards use in a compression setting. This enables us to produce convincing reconstructions for very low bitrates.


\section{Background \& Related Work}
\label{sec:background}

\vspace{-0.05in}
\subsection{Traditional compression techniques}
\label{sec:traditional}
\vspace{-0.05in}

Compression, in general, is very closely related to pattern recognition. If we are able to discover structure in our input, we can eliminate this redundancy to represent it more succinctly. In traditional codecs such as JPEG and JP2, this is achieved via a pipeline which roughly breaks down into 3 modules: transformation, quantization, and encoding (\citet{wallace1992jpeg} and \citet{rabbani2002overview} provide great overviews of the JPEG standards). 

In traditional codecs, since all components are hard-coded, they are heavily engineered to fit together. For example, the coding scheme is custom-tailored to match the distribution of the outputs of the preceding transformation. JPEG, for instance, employs $8\times 8$ block DCT transforms, followed by run-length encoding which exploits the sparsity pattern of the resultant frequency coefficients. JP2 employs an adaptive arithmetic coder to capture the distribution of coefficient magnitudes produced by the preceding multi-resolution wavelet transform.

However, despite the careful construction and assembly of these pipelines, there still remains significant room for improvement of compression efficiency. For example, the transformation is fixed in place irrespective of the distribution of the inputs, and is not adapted to their statistics in any way. In addition, hard-coded approaches often compartmentalize the loss of information within the quantization step. As such, the transformation module is chosen to be bijective: however, this limits the ability to reduce redundancy prior to coding. Moreover, the encode-decode pipeline cannot be optimized for a particular metric beyond manual tweaking: even if we had the perfect metric for image quality assessment, traditional approaches cannot directly optimize their reconstructions for it.

\begin{figure*}[t!]
\centering
\includegraphics[width=\textwidth,trim=0cm 0cm 0cm 0cm,clip]{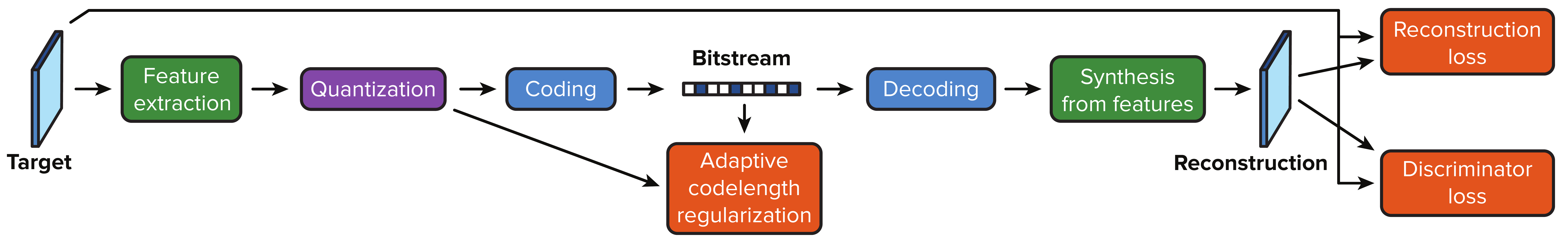}%
\vspace{-0.05in}
\caption{Our overall model architecture. The feature extractor, described in Section \ref{sec:extraction}, discovers structure and reduces redundancy via the pyramidal decomposition and interscale alignment modules. The lossless coding scheme, described in Section \ref{sec:coding}, further compresses the quantized tensor via bitplane decomposition and adaptive arithmetic coding. The adaptive codelength regularization then modulates the expected code length to a prescribed target bitrate. Distortions between the target and its reconstruction are penalized by the reconstruction loss. The discriminator loss, described in Section \ref{sec:gan}, encourages visually pleasing reconstructions by penalizing discrepancies between their distributions and the targets'.}
\label{fig:architecture}
\vspace{-0.1in}
\end{figure*}

\vspace{-0.05in}
\subsection{ML-based lossy image compression}
\label{sec:related}

In approaches based on machine learning, structure is automatically \emph{discovered}, rather than manually engineered. 

One of the first such efforts by \citet{bottou1998high}, for example, introduced the DjVu format for document image compression, which employs techniques such as segmentation and K-means clustering separate foreground from background, and analyze the document's contents. 

At a high level, one natural approach to implement the encoder-decoder image compression pipeline is to use an autoencoder to map the target through a bitrate bottleneck, and train the model to minimize a loss function penalizing it from its reconstruction. This requires carefully constructing a feature extractor and synthesizer for the encoder and decoder, selecting an appropriate objective, and possibly introducing a coding scheme to further compress the fixed-size representation to attain variable-length codes.

Many of the existing ML-based image compression approaches (including ours) follow this general strategy. \citet{toderici2015variable,toderici2016full} explored various transformations for binary feature extraction based on different types of recurrent neural networks; the binary representations were then entropy-coded. \citet{johnston2017improved} enabled another considerable leap in performance by introducing a loss weighted with SSIM \citep{wang2004image}, and spatially-adaptive bit allocation. \citet{twitter2016} and \citet{nyu2016} quantize rather than binarize, and propose strategies to approximate the entropy of the quantized representation: this provides them with a proxy to penalize it. Finally, Pied Piper has recently claimed to employ ML techniques in its Middle-Out algorithm \citep{pied_piper}, although their nature is shrouded in mystery.

\vspace{-0.05in}
\subsection{Generative Adversarial Networks}
\label{sec:gan_background}
\vspace{-0.05in}

One of the most exciting innovations in machine learning in the last few years is the idea of Generative Adversarial Networks (GANs) \citep{goodfellow2014generative}. The idea is to construct a \emph{generator} network $\mathscr{G}_{\bPhi}(\cdot)$ whose goal is to synthesize outputs according to a target distribution $p_{\textrm{true}}$, and a \emph{discriminator} network $\mathscr{D}_{\bTheta}(\cdot)$ whose goal is to distinguish between examples sampled from the ground truth distribution, and ones produced by the generator. This can be expressed concretely in terms of the minimax problem:
\begin{align}
\small
\min_{\bPhi} \max_{\bTheta} \bbE_{\rmbx\sim p_{\textrm{true}}}\log \mathscr{D}_{\bTheta}(\rmbx) + \bbE_{\rmbz\sim p_{\rmbz}}\log\left[1 - \mathscr{D}_{\bTheta}(\mathscr{G}_{\bPhi}(\rmbz))\right]\;.\nonumber
\end{align}
This idea has enabled significant progress in photo-realistic image generation \citep{denton2015deep,radford2015unsupervised,salimans2016improved}, single-image super-resolution \citep{ledig2016photo}, image-to-image conditional translation \citep{isola2016image}, and various other important problems.

The adversarial training framework is particularly relevant to the compression world. In traditional codecs, distortions often take the form of blurriness, pixelation, and so on. These artifacts are unappealing, but are increasingly noticeable as the bitrate is lowered. We propose a multiscale adversarial training model to encourage reconstructions to match the statistics of their ground truth counterparts, resulting in sharp and visually pleasing results even for very low bitrates. As far as we know, we are the first to propose using GANs for image compression. 


\section{Model}
\label{sec:model}
\vspace{-0.05in}
Our model architecture is shown in Figure \ref{fig:architecture}, and comprises a number of components which we briefly outline below. In this section, we limit our focus to operations performed by the encoder: since the decoder simply performs the counterpart inverse operations, we only address exceptions which require particular attention.
\vspace{-0.05in}
\paragraph{Feature extraction.} Images feature a number of different types of structure: across input channels, within individual scales, and across scales. We design our feature extraction architecture to recognize these. It consists of a pyramidal decomposition which analyzes individual scales, followed by an interscale alignment procedure which exploits structure shared across scales.
\vspace{-0.05in}
\paragraph{Code computation and regularization.} This module is responsible for further compressing the extracted features. It quantizes the features, and encodes them via an adaptive arithmetic coding scheme applied on their binary expansions. An adaptive codelength regularization is introduced to penalize the entropy of the features, which the coding scheme exploits to achieve better compression.
\vspace{-0.05in}
\paragraph{Discriminator loss.} We employ adversarial training to pursue realistic reconstructions. We dedicate Section \ref{sec:gan} to describing our GAN formulation.

\subsection{Feature extraction}
\label{sec:extraction}
 
\subsubsection{Pyramidal decomposition}
Our pyramidal decomposition encoder is loosely inspired by the use of wavelets for multiresolution analysis, in which an input is analyzed recursively via feature extraction and downsampling operators \citep{Mallat:1989:TMS:67253.67254}. The JPEG~2000 standard, for example, employs discrete wavelet transforms with the Daubechies 9/7 kernels \citep{DBLP:journals/tip/AntoniniBMD92,rabbani2002overview}. This transform is in fact a linear operator, which can be entirely expressed via compositions of convolutions with only two \emph{hard-coded and separable} $9\times 9$ filters applied irrespective of scale, and independently for each channel. 

The idea of a pyramidal decomposition has been employed in machine learning: for instance, \citet{mathieu2015deep} uses a pyramidal composition for next frame prediction, and \citet{denton2015deep} uses it for image generation. The spectral representations of CNN activations have also been investigated by \citet{rippel2015spectral} to enable processing across a spectrum of scales, but this approach does not enable FIR processing as does wavelet analysis.

\begin{figure}[b!]
\centering
\includegraphics[width=\columnwidth,trim=0cm 0cm 0cm 0cm,clip]{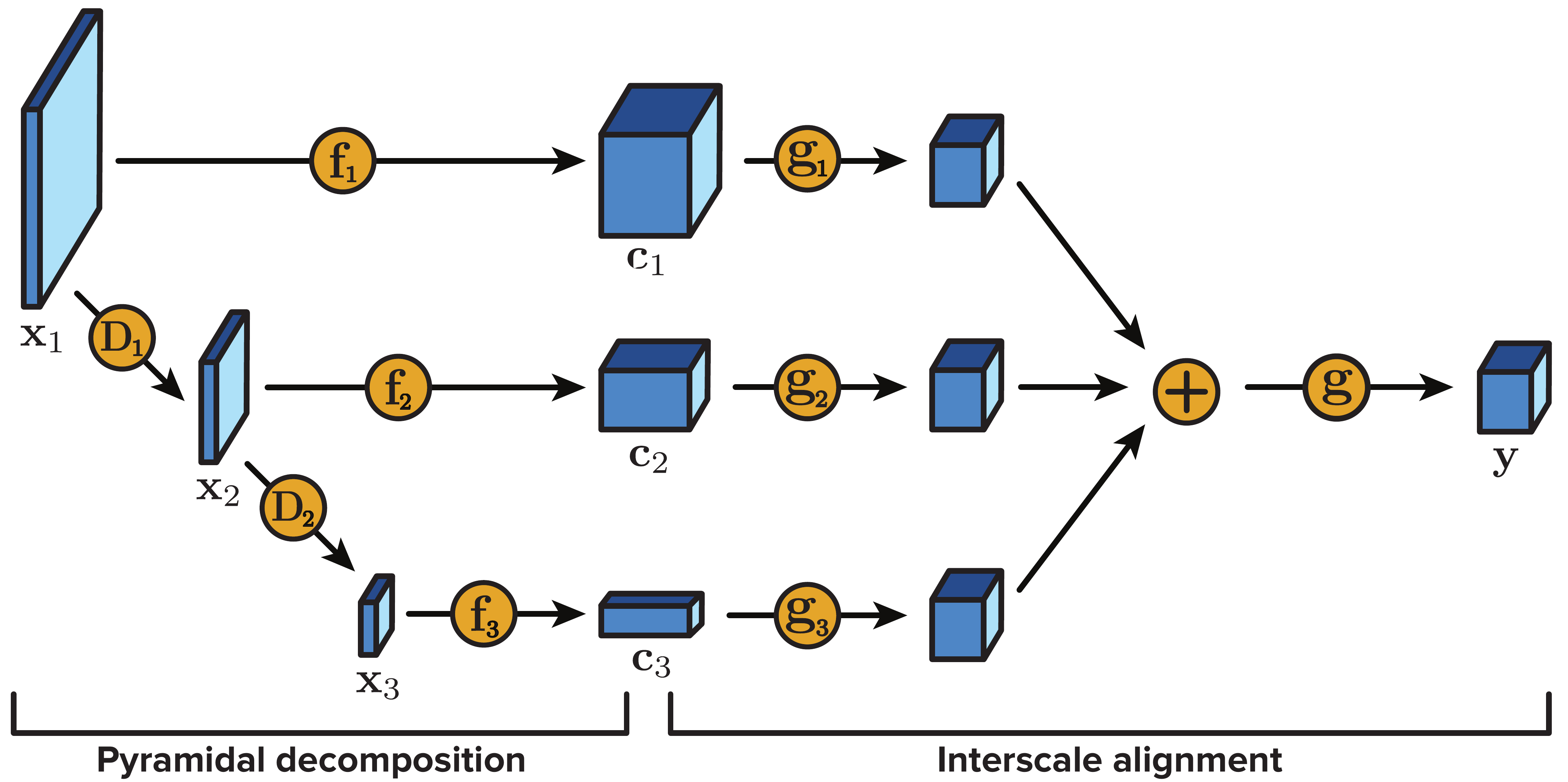}%
\caption{The coefficient extraction pipeline, illustrated for 3 scales. The pyramidal decomposition module discovers structure within individual scales. The extracted coefficient maps are then aligned to discover joint structure across the different scales.}
\label{fig:feature_extraction}
\end{figure}

We generalize the wavelet decomposition idea to learn optimal, nonlinear extractors individually for each scale. Let us assume an input $\rmbx$ to the model, and a total of $M$ scales. We perform recursive analysis: let us denote $\rmbx_m$ as the input to scale $m$; we set the input to the first scale $\rmbx_1=\rmbx$ as the input to the model. For each scale $m$, we perform two operations: first, we extract coefficients $\rmbc_m=\rmbf_m(\rmbx_m)\in\reals^{C_m\times H_m\times W_m}$ via some parametrized function $\rmbf_m(\cdot)$ for output channels $C_m$, height $H_m$ and width $W_m$. Second, we compute the input to the next scale as $\rmbx_{m+1}=\rmbD_m(\rmbx_m)$ where $\rmbD_m(\cdot)$ is some downsampling operator (either fixed or learned). 

Our pyramidal decomposition architecture is illustrated in Figure \ref{fig:feature_extraction}. In practice, we extract across a total of $M = 6$ scales. The feature extractors for the individual scales are composed of a sequence of convolutions with kernels $3\times 3$ or $1\times 1$ and ReLUs with a leak of $0.2$. We learn all downsamplers as $4\times 4$ convolutions with a stride of 2.

\subsubsection{Interscale alignment}
\label{sec:alignment}
Interscale alignment is designed to leverage information shared across different scales --- a benefit not offered by the classic wavelet analysis. It takes in as input the set of coefficients extracted from the different scales $\{\rmbc_m\}_{m=1}^M\subset\reals^{C_m\times H_m\times W_m}$, and produces a single tensor of a target output dimensionality $C\times H\times W$. 

To do this, we first map each input tensor $\rmbc_m$ to the target dimensionality via some parametrized function $\rmbg_m(\cdot)$. This involves ensuring that this function spatially resamples $\rmbc_m$ to the appropriate output map size $H\times W$, and outputs the appropriate number of channels $C$. We then sum $\rmbg_m(\rmbc_m), m = 1, \ldots, M$, and apply another parametrized non-linear transformation $\rmbg(\cdot)$ for joint processing. 

The interscale alignment module can be seen in Figure \ref{fig:feature_extraction}. We denote its output as $\rmby$. In practice, we choose each $\rmbg_m(\cdot)$ as a convolution or a deconvolution with an appropriate stride to produce the target spatial map size $H\times W$; see Section \ref{sec:setup} for a more detailed discussion. We choose $\rmbg(\cdot)$ simply as a sequence of $3\times 3$ convolutions.


\subsection{Code computation and regularization}
\label{sec:coding}
Given the extracted tensor $\rmby\in\reals^{C\times H\times W}$, we proceed to quantize it and encode it. This pipeline involves a number of components which we overview here and describe in detail throughout this section.

\paragraph{Quantization.} The tensor $\rmby$ is quantized to bit precision~$B$: 
\begin{align}
\hat{\rmby} := \textsc{Quantize}_B(\rmby)\;.\nonumber
\end{align}

\paragraph{Bitplane decomposition.} The quantized tensor $\hat{\rmby}$ is transformed into a binary tensor suitable for encoding via a lossless bitplane decomposition:
\begin{align}
\rmbb := \textsc{BitplaneDecompose}_B(\hat{\rmby})\in\{0, 1\}^{B\times C\times H\times W}\;.\nonumber
\end{align}

\paragraph{Adaptive arithmetic coding.} The adaptive arithmetic coder (AAC) is trained to leverage the structure remaining in the data. It encodes $\rmbb$ into its final variable-length binary sequence $\rmbs$ of length $\ell(\rmbs)$:
\begin{align}
\rmbs := \textsc{AACEncode}(\rmbb)\in\{0, 1\}^{\ell(\rmbs)}\;.\nonumber
\end{align}
\paragraph{Adaptive codelength regularization.} The adaptive codelength regularization (ACR) modulates the distribution of the quantized representation $\hat{\rmby}$ to achieve a target expected bit count across inputs:
\begin{align}
\bbE_{\rmbx}[\ell(\rmbs)] \quad\longrightarrow\quad \ell_{\textrm{target}}\;.\nonumber
\end{align}

\subsubsection{Quantization}
Given a desired precision of $B$ bits, we quantize our feature tensor $\rmby$ into $2^B$ equal-sized bins as
\begin{align}
\hy_{chw}:=\textsc{Quantize}_B(y_{chw})=\frac{1}{2^{B-1}}\left\lceil 2^{B-1}y_{chw}\right\rceil\;.\nonumber
\end{align}
For the special case $B = 1$, this reduces exactly to a binary quantization scheme. While some ML-based approaches to compression employ such thresholding, we found better performance with the smoother quantization described. We quantize with $B = 6$ for all models in this paper.

\subsubsection{Bitplane decomposition}
\label{sec:bitplane_decomposition}
\begin{figure}[b!]
\centering
\includegraphics[width=\columnwidth,trim=0cm 0cm 0cm 0cm,clip]{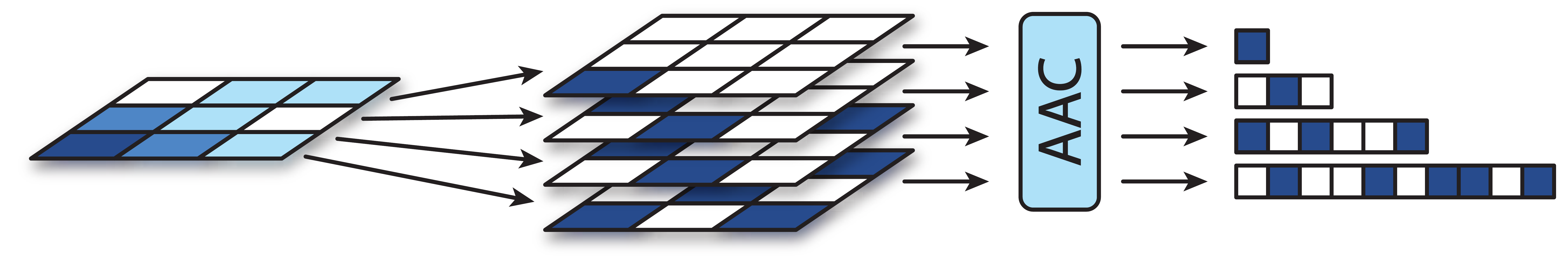}%
\caption{Each of the $C$ spatial maps $\hat{\rmby}_c\in\reals^{H\times W}$ of $\hat{\rmby}$ is decomposed into $B$ bitplanes as each element $\hat{y}_{chw}$ is expressed in its binary expansion. Each set of bitplanes is then fed to the adaptive arithmetic coder for variable-length encoding. The adaptive codelength regularization enables more compact codes for higher bitplanes by encouraging them to feature higher sparsity.}
\label{fig:bitplane_decomposition}
\end{figure}

We decompose $\hat{\rmby}$ into bitplanes. This transformation maps each value $\hat{y}_{chw}$ into its binary expansion of $B$ bits. Hence, each of the $C$ spatial maps $\hat{\rmby}_c\in\reals^{H\times W}$ of $\hat{\rmby}$ expands into $B$ binary \emph{bitplanes}. We illustrate this transformation in Figure \ref{fig:bitplane_decomposition}, and denote its output as $\rmbb\in\{0, 1\}^{B\times C\times H\times W}$. This transformation is lossless. 

As described in Section \ref{sec:entropy_coding}, this decomposition will enable our entropy coder to exploit structure in the distribution of the activations in $\rmby$ to achieve a compact representation. In Section \ref{sec:regularization}, we introduce a strategy to encourage such exploitable structure to be featured.

\subsubsection{Adaptive arithmetic coding}
\label{sec:entropy_coding}
The output $\rmbb$ of the bitplane decomposition is a binary tensor, which contains significant structure: for example, higher bitplanes are sparser, and spatially neighboring bits often have the same value (in Section \ref{sec:regularization} we propose a technique to guarantee presence of these properties). We exploit this low entropy by lossless compression via adaptive arithmetic coding. 

Namely, we associate each bit location in $\rmbb$ with a \emph{context}, which comprises a set of features indicative of the bit value. These are based on the position of the bit as well as the values of neighboring bits. We train a classifier to predict the value of each bit from its context features, and then use these probabilities to compress $\rmbb$ via arithmetic coding.

During decoding, we decompress the code by performing the inverse operation. Namely, we interleave between computing the context of a particular bit using the values of previously decoded bits, and using this context to retrieve the activation probability of the bit and decode it. We note that this constrains the context of each bit to only include features composed of bits already decoded.

\begin{figure*}[t]
\centering
\includegraphics[height=0.085in,trim=0cm 0cm 0cm 0cm,clip]{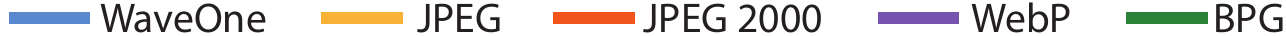}\\
\vspace{0.05in}
\includegraphics[height=1.7in,trim=0cm 0cm 0cm 0cm,clip]{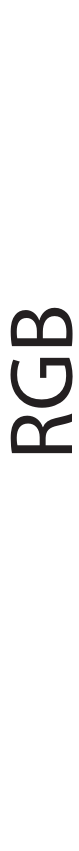}\,\,\,
\includegraphics[height=1.7in,trim=0cm 0cm 0cm 0cm,clip]{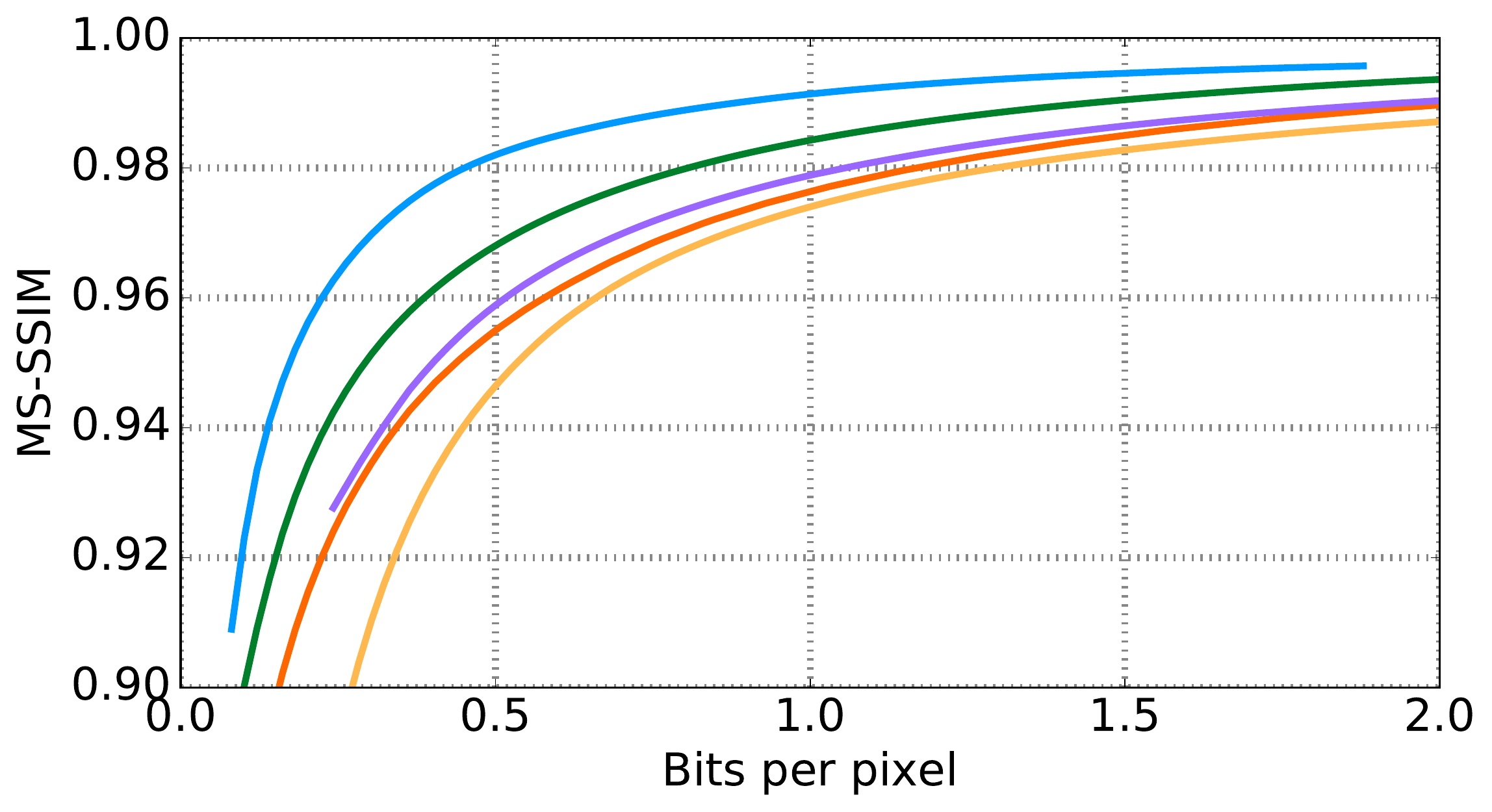}
\includegraphics[height=1.7in,trim=0cm 0cm 0cm 0cm,clip]{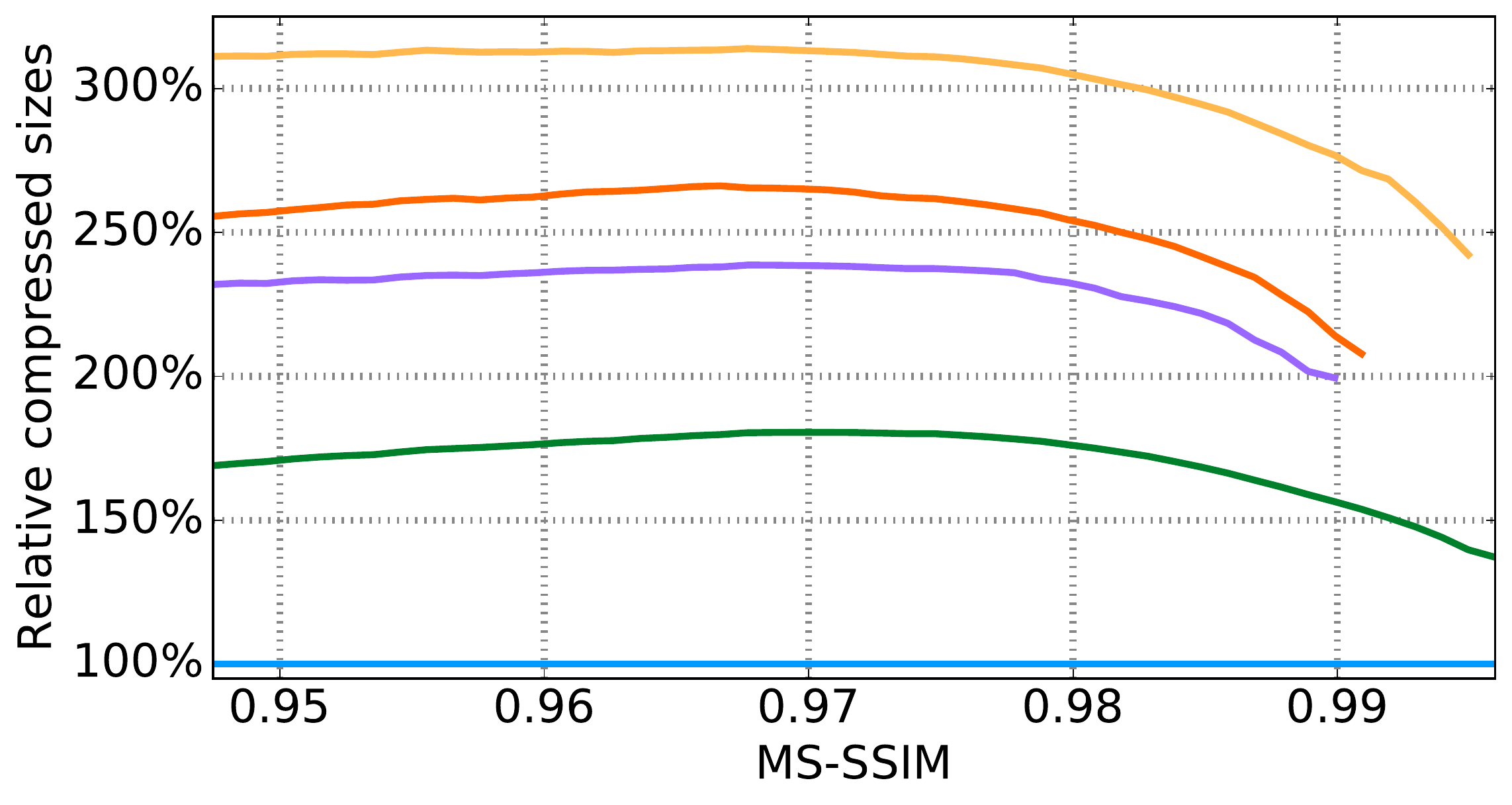} \\
\includegraphics[height=1.7in,trim=0cm 0cm 0cm 0cm,clip]{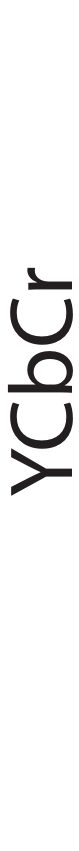}\,\,\,
\includegraphics[height=1.7in,trim=0cm 0cm 0cm 0cm,clip]{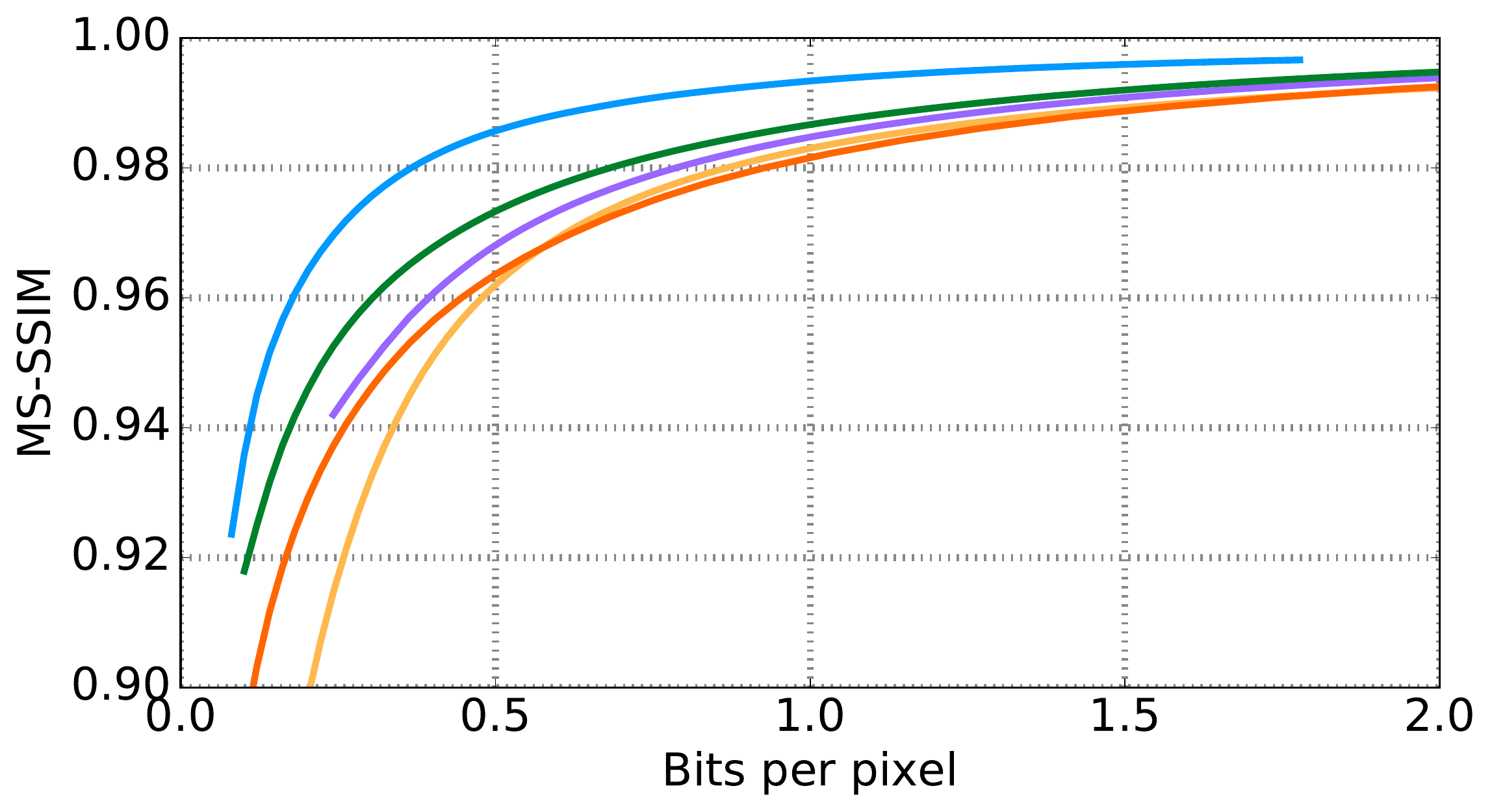}
\includegraphics[height=1.7in,trim=0cm 0cm 0cm 0cm,clip]{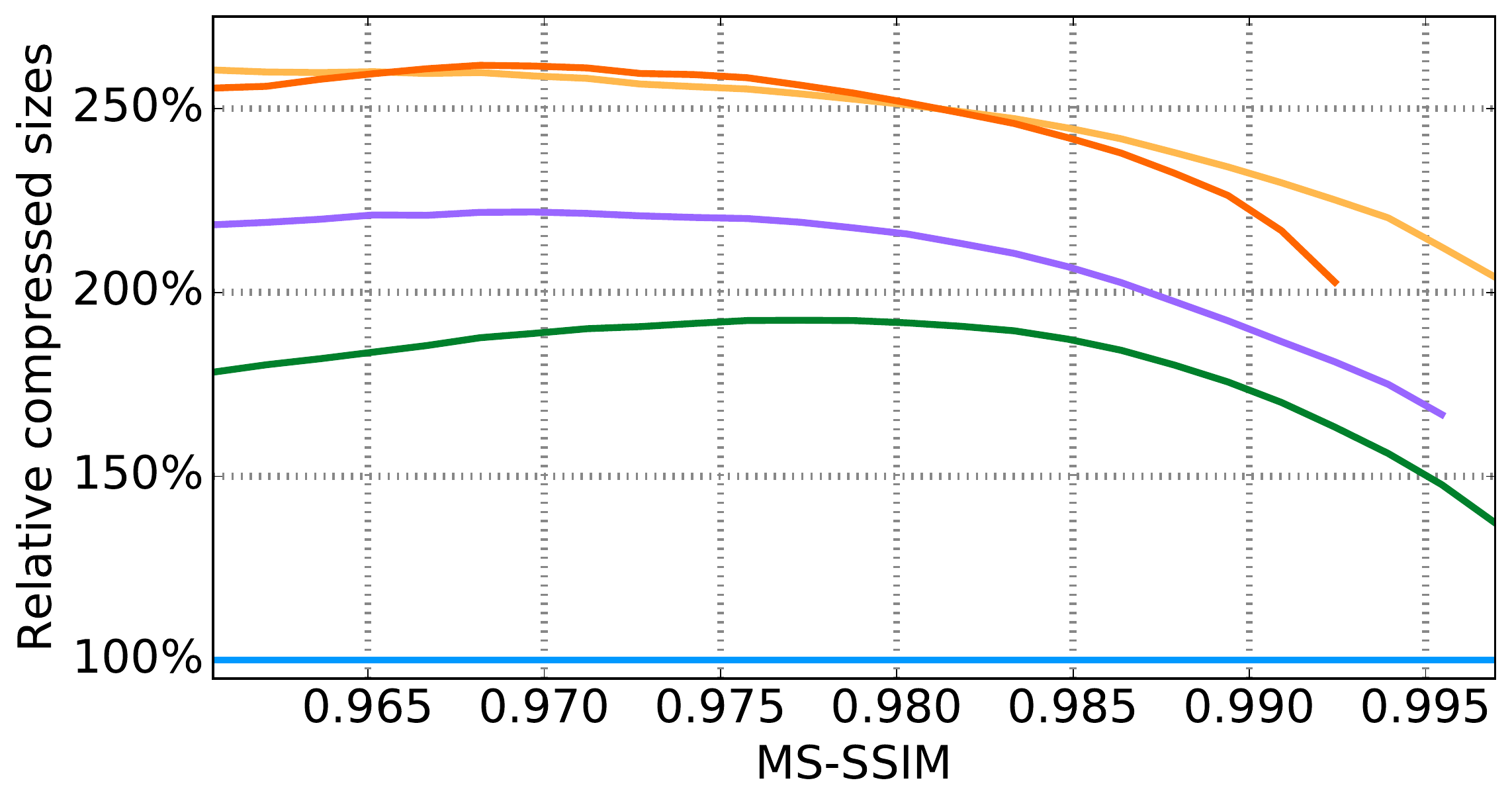}
\vspace{-0.1in}
\caption{Compression results for the RAISE-1k $512\times 768$ dataset, measured in the RGB domain (top row) and YCbCr domain (bottom row). We compare against commercial codecs JPEG, JPEG~2000, WebP and BPG\footref{bpg} (4:2:0 for YCbCr and 4:4:4 for RGB). The plots on the left present average reconstruction quality, as function of the number of bits per pixel fixed for each image. The plots on the right show average compressed file sizes relative to ours for different target MS-SSIM values for each image. In Section \ref{sec:performance} we discuss the curve generation procedures in detail.}
\label{fig:results_raise}
\end{figure*}

\subsubsection{Adaptive codelength regularization}
\label{sec:regularization}

One problem with classic autoencoder architectures is that their bottleneck has fixed capacity. The bottleneck may be too small to represent complex patterns well, which affects quality, and it may be too large for simple patterns, which results in inefficient compression. What we need is a model capable of generating long representations for complex patterns and short for simple ones, while maintaining an expected codelength target over large number of examples. To achieve this, the AAC is necessary, but not sufficient.

We extend the architecture by increasing the dimensionality of $\rmbb$ --- but at the same time controlling its information content, thereby resulting in shorter compressed code $\rmbs = \textsc{AACEncode}(\rmbb)\in\{0, 1\}$. Specifically, we introduce the adaptive codelength regularization (ACR), which enables us to regulate the expected codelength $\bbE_{\rmbx}[\ell(\rmbs)]$ to a target value $\ell_{\textrm{target}}$. This penalty is designed to encourage structure exactly where the AAC is able to exploit it. Namely, we regularize our quantized tensor $\hat{\rmby}$ with
\begin{align}
\mathscr{P}(\hat{\rmby}) &= \frac{\alpha_t}{CHW}\sum_{chw}\Big\{\log_2\left| \hy_{chw}\right|\nonumber \\
&+ \sum_{(x, y)\in S}\log_2\left| \hy_{chw} - \hy_{c(h - y)(w - x)} \right|\Big\}\;,\nonumber
\end{align}
for iteration $t$ and difference index set $S=\{(0, 1), (1, 0), (1, 1), (-1, 1)\}$. The first term penalizes the magnitude of each tensor element, and the second penalizes deviations between spatial neighbors. These enable better prediction by the AAC.

As we train our model, we continuously modulate the scalar coefficient $\alpha_t$ to pursue our target codelength. We do this via a feedback loop. We use the AAC to monitor the mean number of effective bits. If it is too high, we increase $\alpha_t$; if too low, we decrease it. In practice, the model reaches an equilibrium in a few hundred iterations, and is able to maintain it throughout training.

Hence, we get a knob to tune: the ratio of total bits, namely the $BCHW$ bits available in $\rmbb$, to the target number of effective bits $\ell_{\textrm{target}}$. This allows exploring the trade-off of increasing the number of channels or spatial map size of $\rmbb$ at the cost of increasing sparsity. We find that a  total-to-target ratio of $BCHW/\ell_{\textrm{target}} = 4$ works well across all architectures we have explored.


\begin{figure*}[t]
\centering
\includegraphics[height=0.085in,trim=0cm 0cm 0cm 0cm,clip]{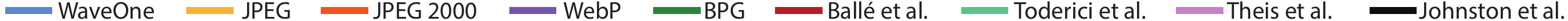}\\
\vspace{0.05in}
\includegraphics[height=1.7in,trim=0cm 0cm 0cm 0cm,clip]{./figures/label_rgb.pdf}\,\,\,
\includegraphics[height=1.7in,trim=0cm 0cm 0cm 0cm,clip]{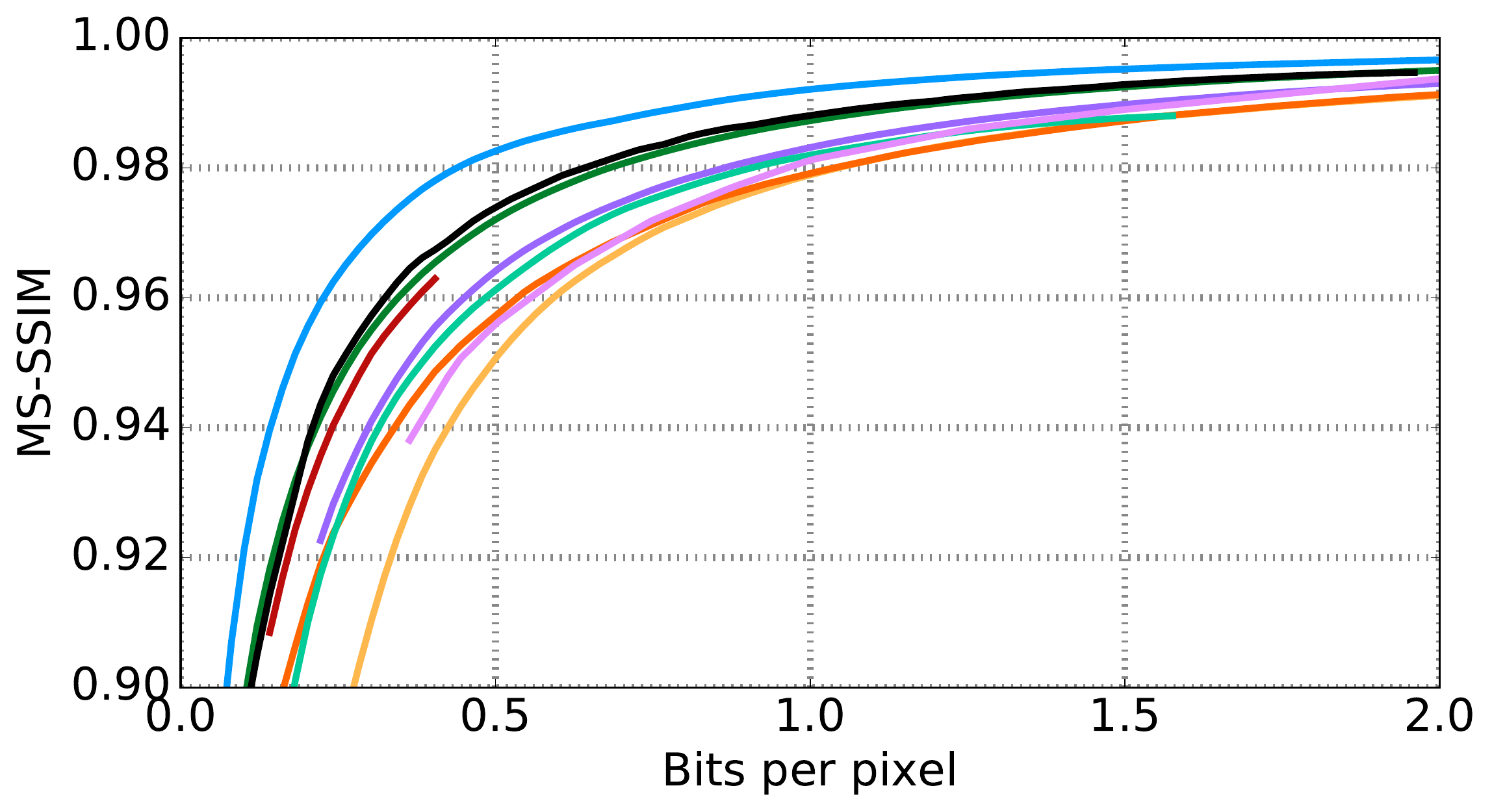}
\includegraphics[height=1.7in,trim=0cm 0cm 0cm 0cm,clip]{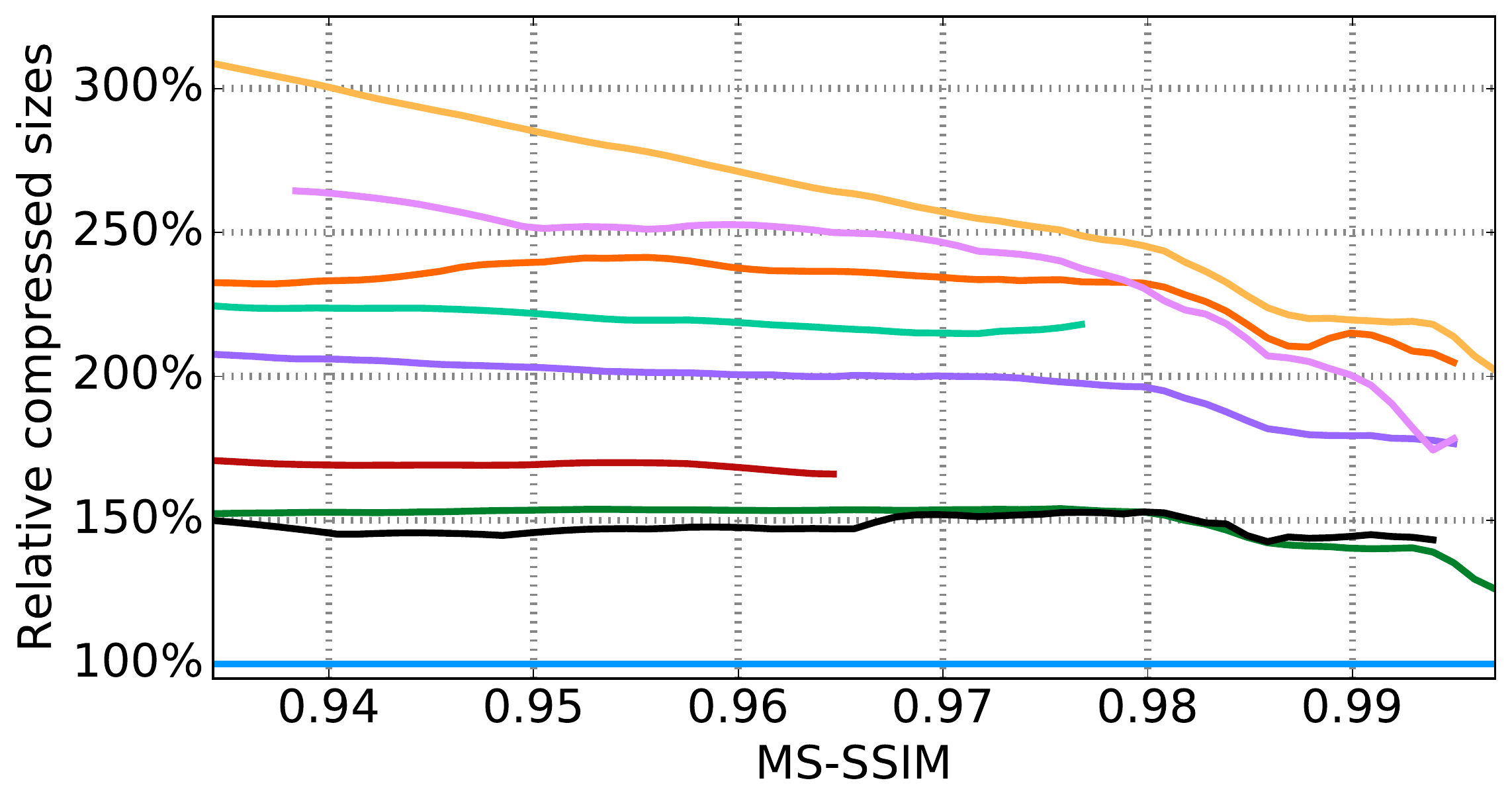} \\
\includegraphics[height=1.7in,trim=0cm 0cm 0cm 0cm,clip]{./figures/label_ycbcr.pdf}\,\,\,
\includegraphics[height=1.7in,trim=0cm 0cm 0cm 0cm,clip]{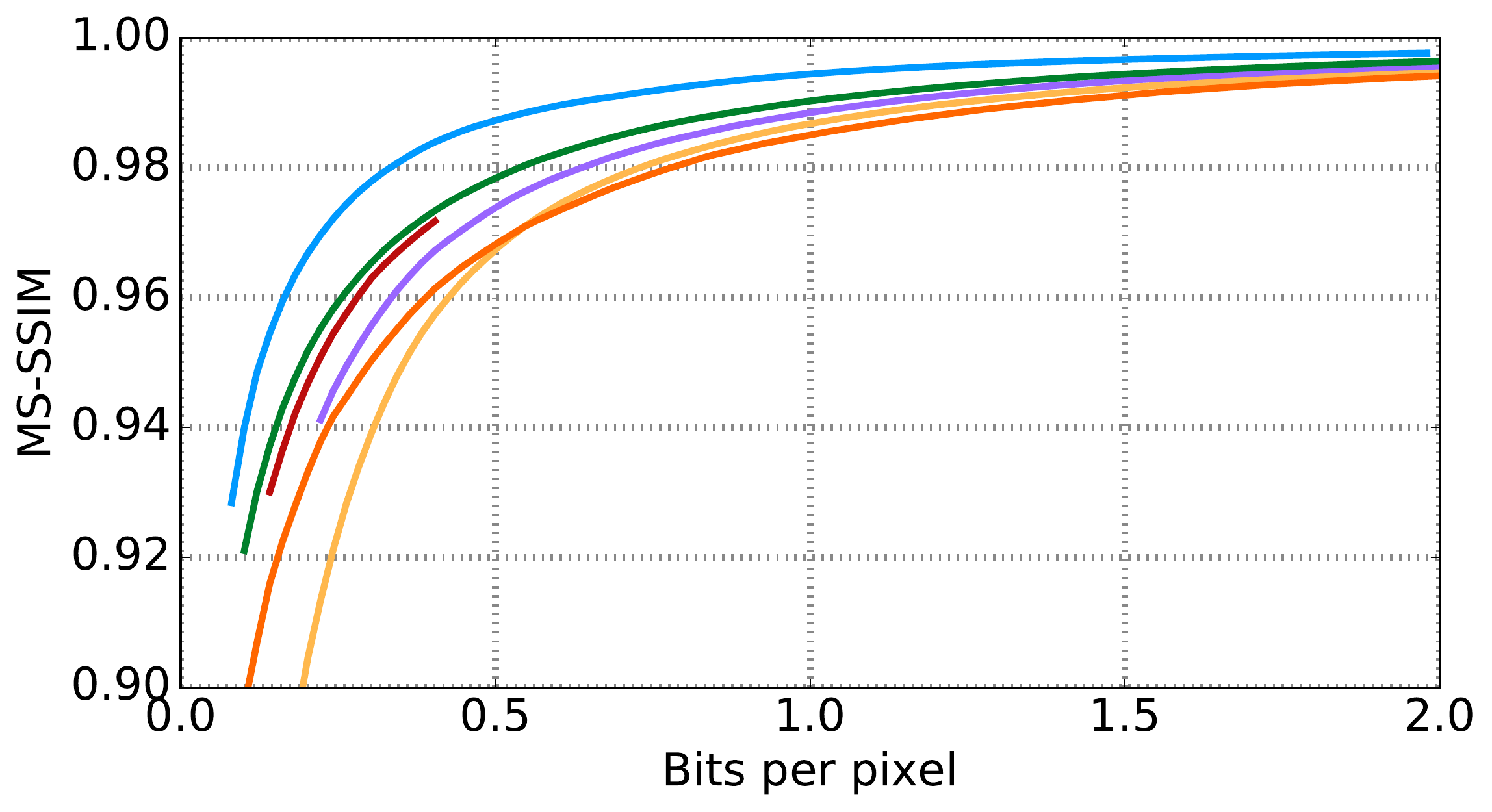}
\includegraphics[height=1.7in,trim=0cm 0cm 0cm 0cm,clip]{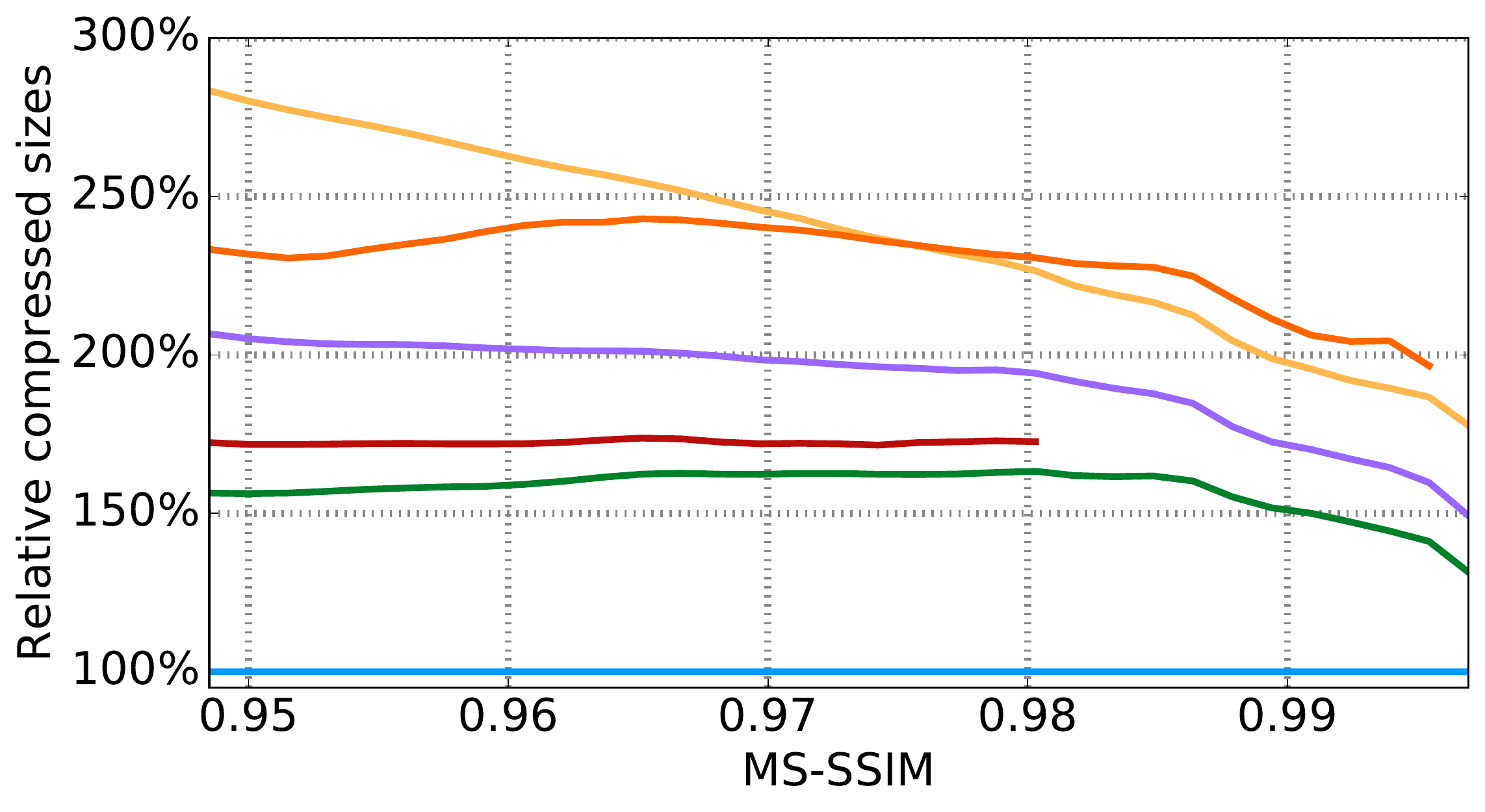}
\vspace{-0.1in}
\caption{Performance on the Kodak PhotoCD dataset measured in the RGB domain (top row) and YCbCr domain (bottom row). We compare against commercial codecs JPEG, JPEG~2000, WebP and BPG\footref{bpg} (4:2:0 for YCbCr and 4:4:4 for RGB), as well as recent ML-based compression work by \citet{toderici2016full}\footref{google}, \citet{twitter2016}\footref{twitter}, \citet{nyu2016}\footref{nyu}, and \citet{johnston2017improved}\footref{twitter} in all settings where results exist. The plots on the left present average reconstruction quality, as function of the number of bits per pixel fixed for each image. The plots on the right show average compressed file sizes relative to ours for different target MS-SSIM values for each image.}
\label{fig:results_kodak}
\end{figure*}

\section{Realistic Reconstructions via Multiscale Adversarial Training}
\label{sec:gan}

\subsection{Discriminator design}
In our compression approach, we take the generator as the encoder-decoder pipeline, to which we append a discriminator --- albeit with a few key differences from existing GAN formulations. 

In many GAN approaches featuring both a reconstruction and a discrimination loss, the target and the reconstruction are treated independently: each is separately assigned a label indicating whether it is real or fake. In our formulation, we consider the target and its reconstruction jointly as a single example: we compare the two by asking \emph{which} of the two images is the real one. 

To do this, we first swap between the target and reconstruction in each input pair to the discriminator with uniform probability. Following the random swap, we propagate each set of examples through the network. However, instead of producing an output for classification at the very last layer of the pipeline, we accumulate scalar outputs along branches constructed along it at different depths. We average these to attain the final value provided to the terminal sigmoid function. This multiscale architecture allows aggregating information across different scales, and is motivated by the observation that undesirable artifacts vary as function of the scale in which they are exhibited. For example, high-frequency artifacts such as noise and blurriness are discovered by earlier scales, whereas more abstract discrepancies are found in deeper scales. 

We apply our discriminator $\mathscr{D}_{\bTheta}$ on the aggregate sum across scales, and proceed to formulate our objectives as described in Section \ref{sec:gan_background}. The complete discriminator architecture is illustrated in Figure \ref{fig:disc_architecture}. 

\subsection{Adversarial training}
Training a GAN system can be tricky due to optimization instability. In our case, we were able to address this by designing a training scheme adaptive in two ways. First, the reconstructor is trained by both the confusion signal gradient as well as the reconstruction loss gradient: we balance the two as function of their gradient magnitudes. Second, at any point during training, we either train the discriminator or propagate confusion signal through the reconstructor, as function of the prediction accuracy of the discriminator. 

More concretely, given lower and upper accuracy bounds $L, U\in[0, 1]$ and discriminator accuracy $a(\mathscr{D}_{\bTheta})$, we apply the following procedure:
\begin{itemize}
\item If $a < L$: freeze propagation of confusion signal through the reconstructor, and train the discriminator.
\item If $L \leq a < U$: alternate between propagating confusion signal and training the disciminator.
\item If $U \leq a$: propagate confusion signal through the reconstructor, and freeze the discriminator.
\end{itemize}
In practice we used $L = 0.8, U = 0.95$. We compute the accuracy $a$ as a running average over mini-batches with a momentum of $0.8$.

\section{Results}
\label{sec:results}

\subsection{Experimental setup}
\label{sec:setup}
\paragraph{Similarity metric.} We trained and tested all models on the Multi-Scale Structural Similarity Index Metric (MS-SSIM) \cite{wang2003multiscale}. This metric has been specifically designed to match the human visual system, and has been established to be significantly more representative than losses in the $\ell_p$ family and variants such as PSNR.

\paragraph{Color space.} Since the human visual system is much more sensitive to variations in brightness than color, most codecs represent colors in the YCbCr color space to devote more bandwidth towards encoding luma rather than chroma. In quantifying image similarity, then, it is common to assign the Y, Cb, Cr components weights $6/8, 1/8, 1/8$. While many ML-based compression papers evaluate similarity in the RGB space with equal color weights, this does not allow fair comparison with standard codecs such as JPEG, JPEG~2000 and WebP, since they have not been designed to perform optimally in this domain. In this work, we provide comparisons with both traditional and ML-based codecs, and present results in both the RGB domain with equal color weights, as well as in YCbCr with weights as above.

\paragraph{Reported performance metrics.} We present both compression performance of our algorithm, but also its runtime. While the requirement of running the approach in real-time severely constrains the capacity of the model, it must be met to enable feasible deployment in real-life applications. 

\paragraph{Training and deployment procedure.} We trained and tested all models on a GeForce GTX 980 Ti GPU and a custom codebase. We trained all models on $128\times 128$ patches sampled at random from the Yahoo Flickr Creative Commons 100 Million dataset \citep{thomee2016yfcc100m}. 

We optimized all models with Adam \citep{kingma2014adam}. We used an initial learning rate of $3\times 10^{-4}$, and reduced it twice by a factor of 5 during training. We chose a batch size of 16 and trained each model for a total of 400,000 iterations. We initialized the ACR coefficient as $\alpha_0 = 1$. During runtime we deployed the model on arbitrarily-sized images in a fully-convolutional way. To attain the rate-distortion (RD)curves presented in Section \ref{sec:performance}, we trained models for a range of target bitrates $\ell_{\textrm{target}}$.

\stepcounter{footnote}
\footnotetext{The Kodak PhotoCD dataset can be found at \url{http://r0k.us/graphics/kodak}.\label{kodak}}

\stepcounter{footnote}
\footnotetext{The results of \citet{toderici2016full} on the Kodak RGB dataset are available at \url{http://github.com/tensorflow/models/tree/master/compression}.\label{google}}

\stepcounter{footnote}
\footnotetext{We have no access to reconstructions by \citet{twitter2016} and \citet{johnston2017improved}, so we carefully transcribed their results, only available in RGB, from the graphs in their paper.\label{twitter}}

\stepcounter{footnote}
\footnotetext{Reconstructions by \citet{nyu2016} of images in the Kodak dataset can be found at \url{http://www.cns.nyu.edu/~lcv/iclr2017/} for both RGB and YCbCr and across a spectrum of BPPs. We use these to compute RD curves by the procedure described in this section.\label{nyu}}

\stepcounter{footnote}
\footnotetext{An implementation of the BPG codec is available at \url{http://bellard.org/bpg}.\label{bpg}}

\subsection{Performance}
\label{sec:performance}
\vspace{-0.05in}

We present several types of results:
\begin{enumerate}
    \item Average MS-SSIM as function of the BPP fixed for each image, found in Figures \ref{fig:results_raise} and \ref{fig:results_kodak}, and Table \ref{tab:results}. 
    \item Average compressed file sizes relative to ours as function of the MS-SSIM fixed for each image, found in Figures \ref{fig:results_raise} and \ref{fig:results_kodak}, and Table \ref{tab:results}.
    \item Encode and decode timings as function of MS-SSIM, found in Figure \ref{fig:timings}, in the appendix, and Table \ref{tab:results}.
    \item Visual examples of reconstructions of different compression approaches for the same BPP, found in Figure~\ref{fig:examples} and in the appendix.
\end{enumerate}

\paragraph{Test sets.} To enable comparison with other approaches, we first present performance on the Kodak PhotoCD dataset\footref{kodak}. While the Kodak dataset is very popular for testing compression performance, it contains only 24 images, and hence is susceptible to overfitting and does not necessarily fully capture broader statistics of natural images. As such, we additionally present performance on the RAISE-1k dataset \citep{dang2015raise} which contains 1,000 raw images. We resized each image to size $512\times 768$ (backwards if vertical): we intend to release our preparation code to enable reproduction of the dataset used.

We remark it is important to use a dataset of raw, rather than previously compressed, images for codec evaluation. Compressing an image introduces artifacts with a bias particular to the codec used, which results in a more favorable RD curve if it compressed \emph{again} with the same codec. See Figure \ref{fig:jpeg_recompressions} for a plot demonstrating this effect.

\begin{figure}[t]
\centering
\includegraphics[height=0.085in,trim=0cm 0cm 0cm 0cm,clip]{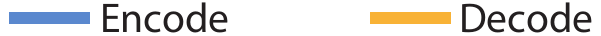}\\
\includegraphics[width=0.9\columnwidth,trim=0cm 0cm 0cm 0cm,clip]{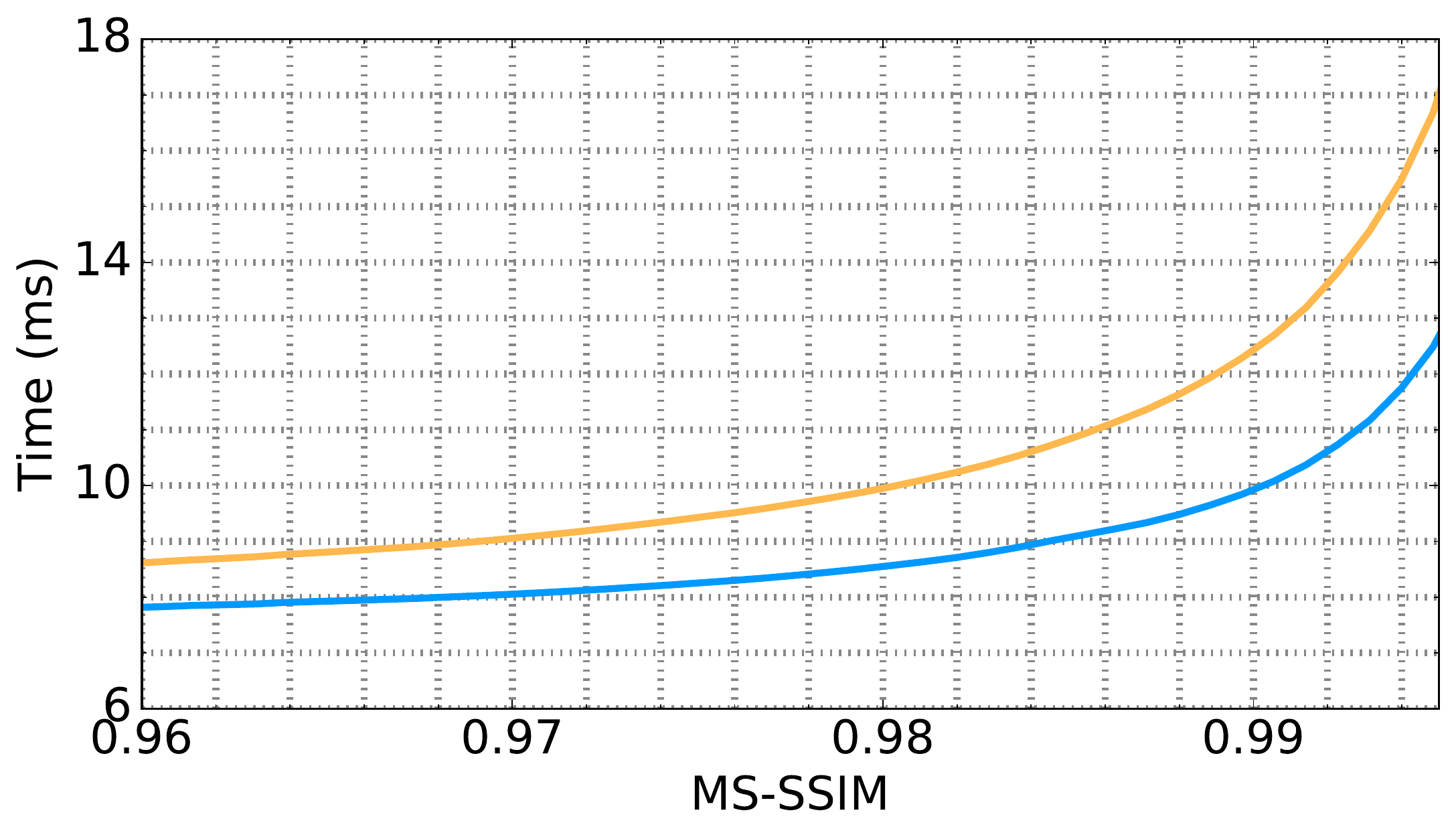}
\vspace{-0.15in}
\caption{Average times to encode and decode images from the RAISE-1k $512\times 768$ dataset using our approach.}
\label{fig:timings}
\end{figure}

\paragraph{Codecs.} We compare against commercial compression techniques JPEG, JPEG~2000, WebP, as well as recent ML-based compression work by \citet{toderici2016full}\footref{google}, \citet{twitter2016}\footref{twitter}, \citet{nyu2016}\footref{nyu}, and \citet{johnston2017improved}\footref{twitter} in all settings in which results are available. We also compare to BPG\footref{bpg} (4:2:0 and 4:4:4) which, while not widely used, surpassed all other codecs in the past. We use the best-performing configuration we can find of JPEG, JPEG~2000, WebP, and BPG, and reduce their bitrates by their respective header lengths for fair comparison.

\paragraph{Performance evaluation.} For each image in each test set, each compression approach, each color space, and for the selection of available compression rates, we recorded (1) the BPP, (2) the MS-SSIM (with components weighted appropriately for the color space), and (3) the computation times for encoding and decoding. 

It is important to take great care in the design of the performance evaluation procedure. Each image has a separate RD curve computed from all available compression rates for a given codec: as \citet{nyu2016} discusses in detail, different summaries of these RD curves lead to disparate results. In our evaluations, to compute a given curve, we sweep across values of the independent variable (such as bitrate). We interpolate each individual RD curve at this independent variable value, and average all the results. To ensure accurate interpolation, we sample densely across rates for each codec.

\paragraph{Acknowledgements} We are grateful to Trevor Darrell, Sven Strohband, Michael Gelbart, Robert Nishihara, Albert Azout, and Vinod Khosla for meaningful discussions and input.

\bibliography{main}
\bibliographystyle{icml2017}

\clearpage
\newpage
\begin{appendices}
\twocolumn[
\icmltitle{Real-Time Adaptive Image Compression: Supplementary Material}
]
\icmltitlerunning{Real-Time Adaptive Image Compression: Supplementary Material}
 \begin{figure}[h!]
 \begin{minipage}{\textwidth}
 \centering
\includegraphics[height=0.085in,trim=0cm 0cm 0cm 0cm,clip]{./figures/legend_short.pdf}\\
\subfigure[Encode times.]{\includegraphics[width=0.35\columnwidth,trim=0cm 0cm 0cm 0cm,clip]{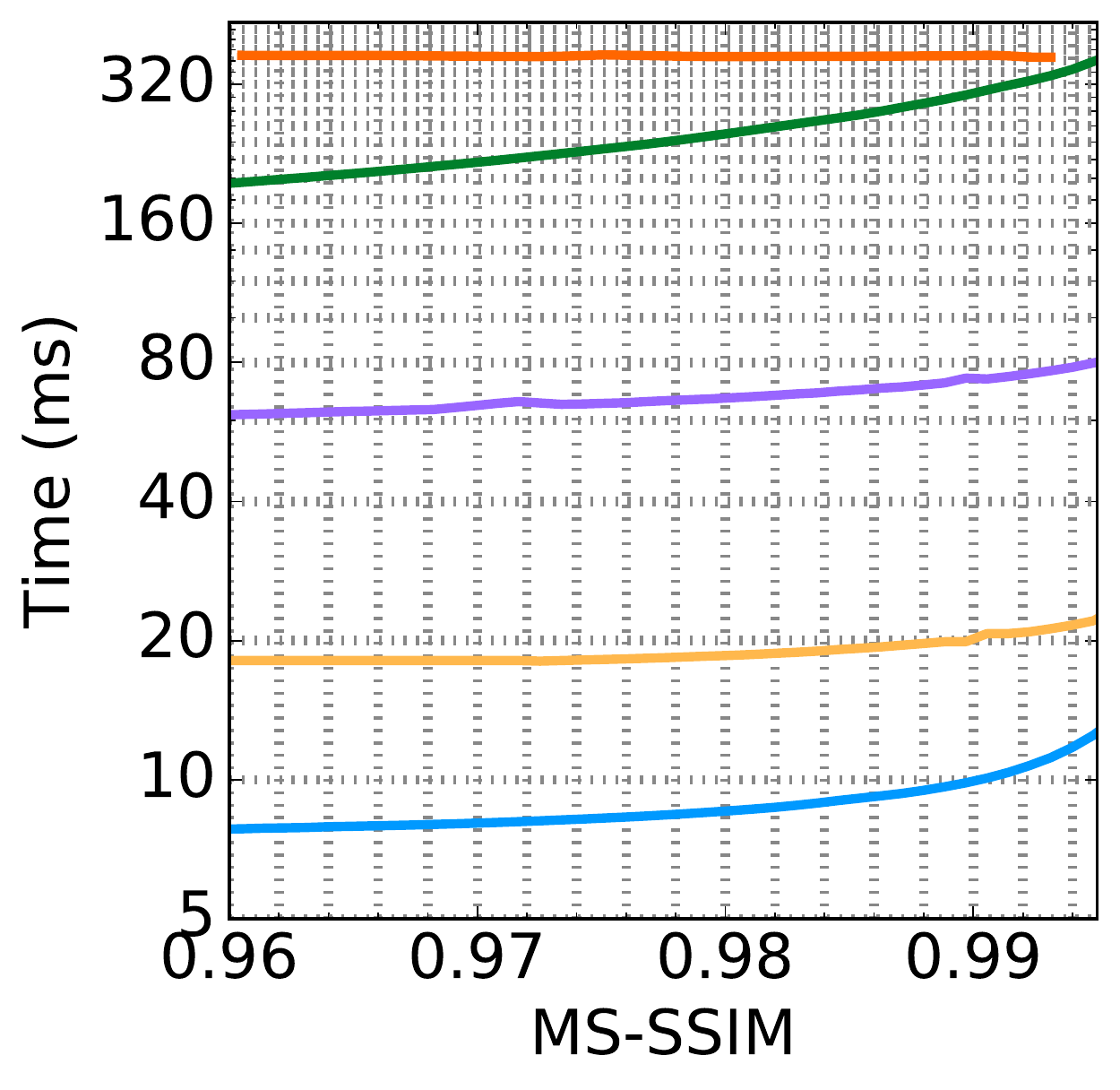}}
\subfigure[Decode times.]{\includegraphics[width=0.35\columnwidth,trim=0cm 0cm 0cm 0cm,clip]{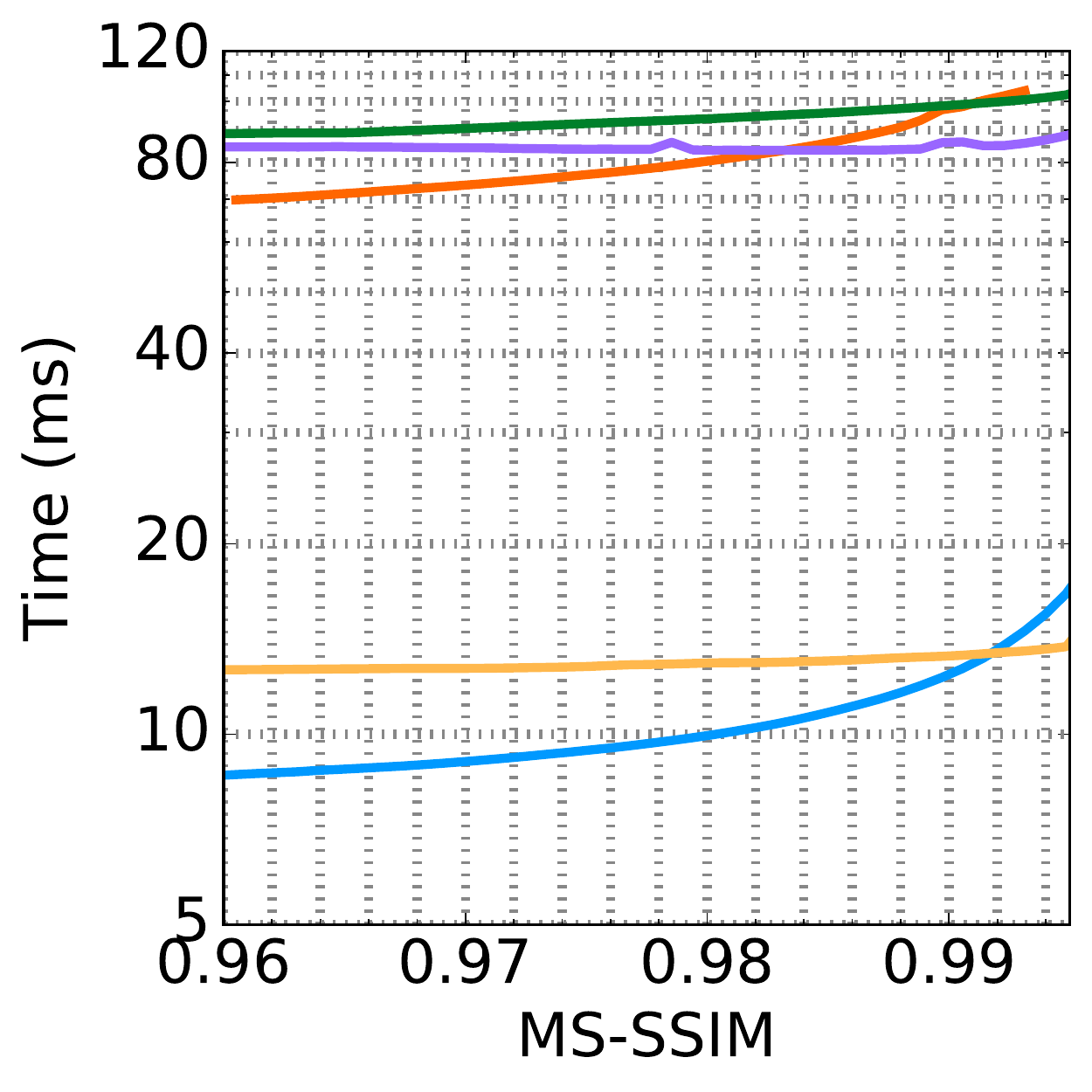}}
\caption{Average times to encode and decode images from the RAISE-1k $512\times 768$ dataset. Note our codec was run on GPU.}
\label{fig:timings_extended}
 \end{minipage}
 \end{figure}

\begin{figure}[h!]
\vspace{3.5in}
\hspace{-3.5in}
\begin{minipage}{\textwidth}
\centering
\includegraphics[width=0.8\textwidth,trim=0cm 0cm 0cm 0cm,clip]{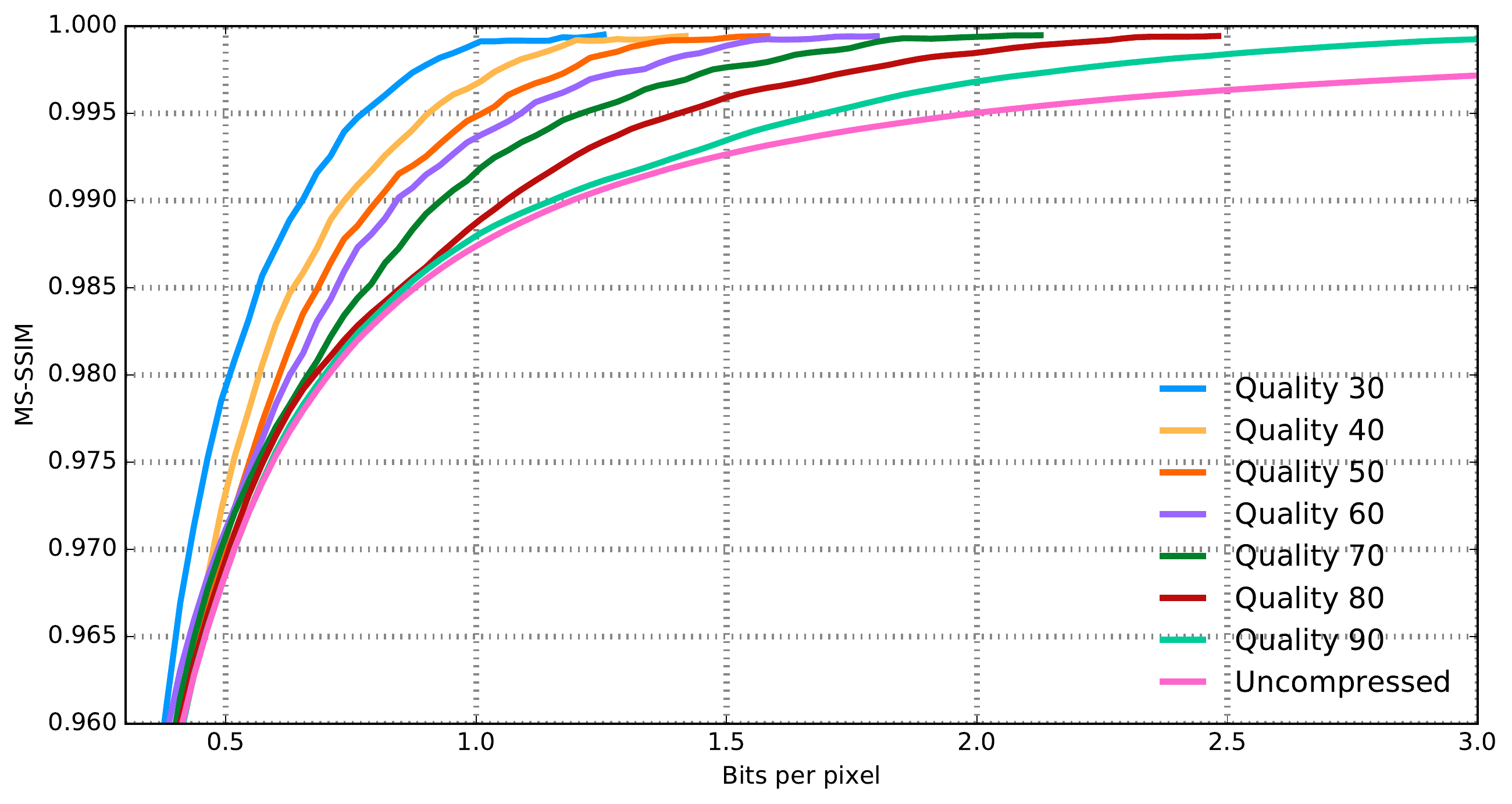}%
\caption{We used JPEG to compress the Kodak dataset at various quality levels. For each, we then use JPEG to recompress the images, and plot the resultant rate-distortion curve. It is evident that the more an image has been previously compressed with JPEG, the better JPEG is able to then recompress it.}
\label{fig:jpeg_recompressions}
\end{minipage}
\end{figure}

\newpage
\begin{figure}[h!]
\begin{minipage}{\textwidth}
\centering
\includegraphics[width=\textwidth,trim=0cm 0cm 0cm 0cm,clip]{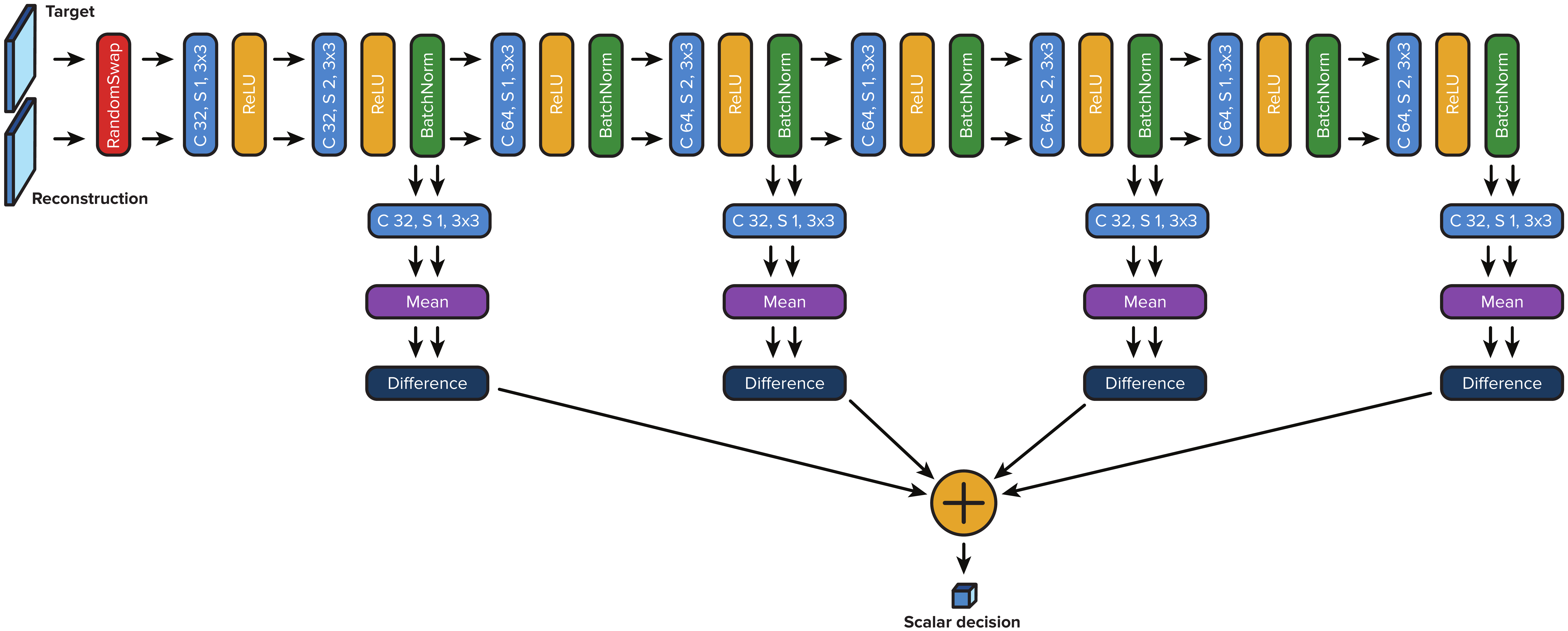}%
\caption{The architecture of the discriminator used in our adversarial training procedure. The first module randomly swaps between the targets and the reconstructions. The goal of the discriminator is to infer which of the two inputs is then the real target, and which is its reconstruction. We accumulate scalar outputs along branches constructed along the processing pipeline, branched out at different depths. We average these to attain the final value provided to the objective sigmoid function. This multiscale architecture allows aggregating information across different scales. In Section \ref{sec:gan} of the main text we discuss the motivation for these architectural choices in more detail.}
\label{fig:disc_architecture}
\end{minipage}
\end{figure}

\newpage
\begin{figure*}[t!]
\vspace{-0.1in}
\begin{widepage}
\begin{multicols}{4}
{\bf\centering JPEG\\ \vspace{0.05in}}
\includegraphics[width=0.25\textwidth,trim=0cm 0cm 0cm 0cm,clip]{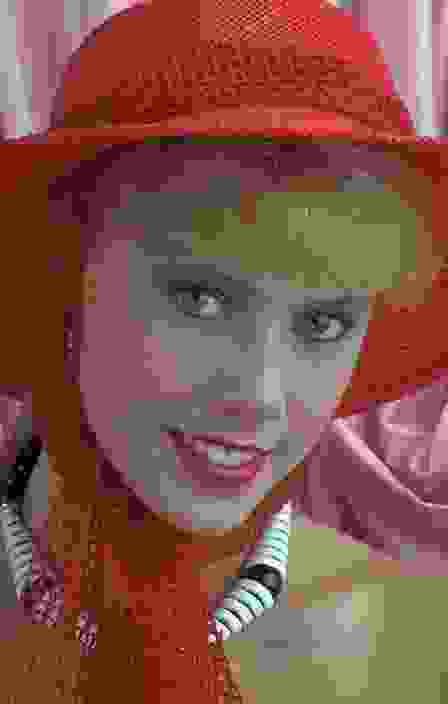}
{\footnotesize\centering 0.0909 BPP\\ \vspace{0.05in}}
\includegraphics[width=0.25\textwidth,trim=0cm 0cm 0cm 0cm,clip]{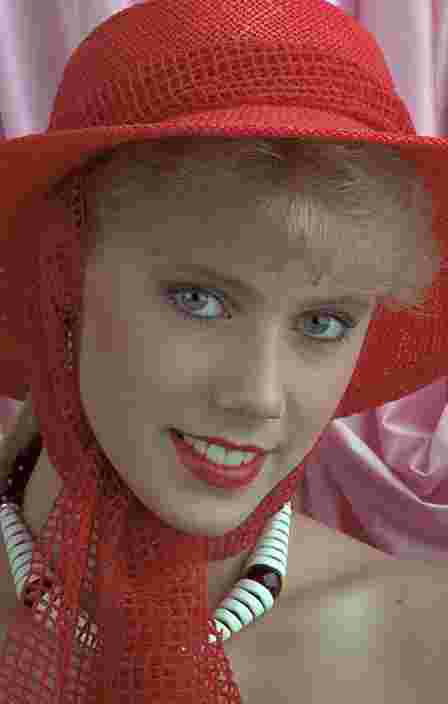}
{\footnotesize\centering 0.1921 BPP\\ \vspace{0.05in}}
\includegraphics[width=0.25\textwidth,trim=0cm 0cm 0cm 0cm,clip]{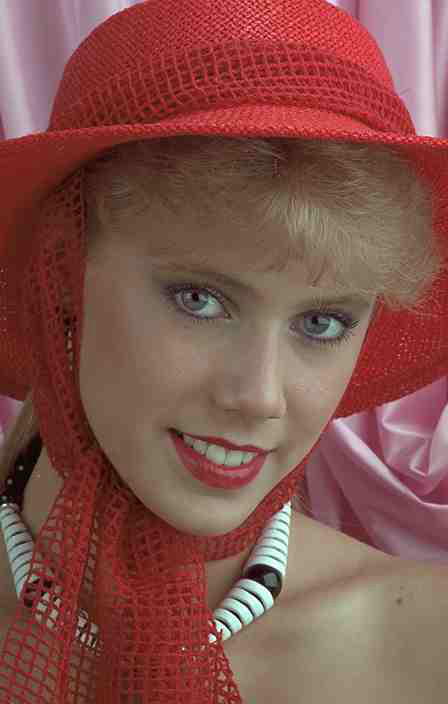}
{\footnotesize\centering 0.4064 BPP\\ \vspace{0.05in}}

{\bf\centering JPEG 2000\\ \vspace{0.05in}}
\includegraphics[width=0.25\textwidth,trim=0cm 0cm 0cm 0cm,clip]{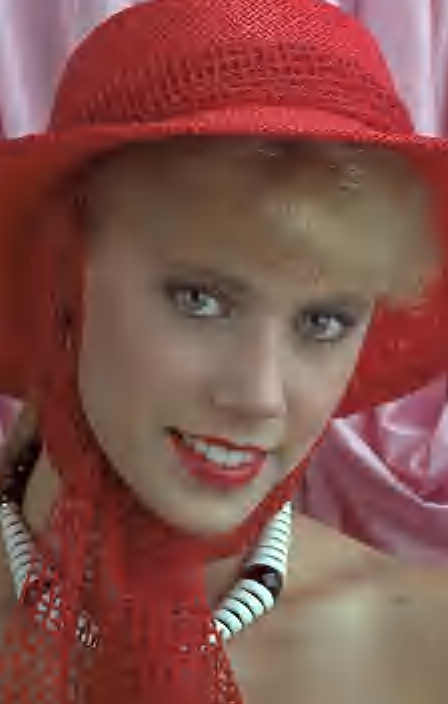}
{\footnotesize\centering 0.0847 BPP\\ \vspace{0.05in}}
\includegraphics[width=0.25\textwidth,trim=0cm 0cm 0cm 0cm,clip]{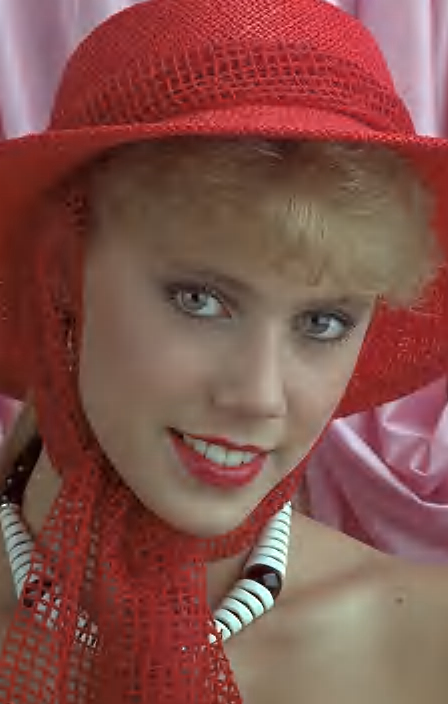}
{\footnotesize\centering 0.1859 BPP\\ \vspace{0.05in}}
\includegraphics[width=0.25\textwidth,trim=0cm 0cm 0cm 0cm,clip]{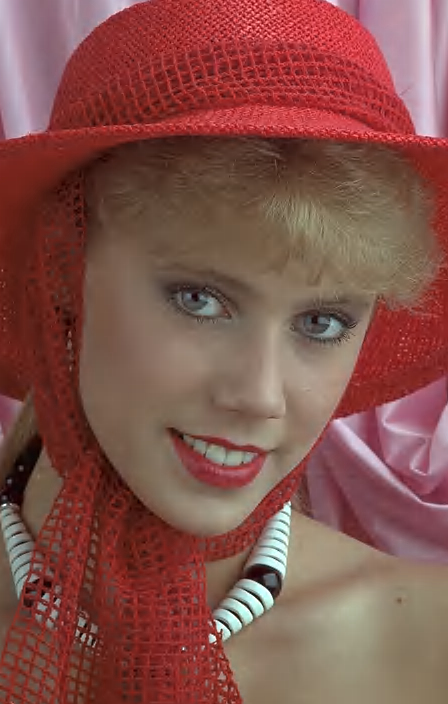}
{\footnotesize\centering 0.4002 BPP\\ \vspace{0.05in}}

{\bf\centering WebP\\ \vspace{0.05in}}
\includegraphics[width=0.25\textwidth,trim=0cm 0cm 0cm 0cm,clip]{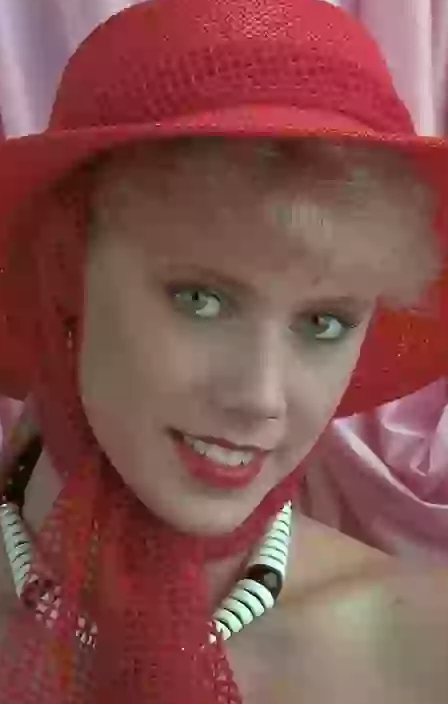}
{\footnotesize\centering 0.1021 BPP\\ \vspace{0.05in}}
\includegraphics[width=0.25\textwidth,trim=0cm 0cm 0cm 0cm,clip]{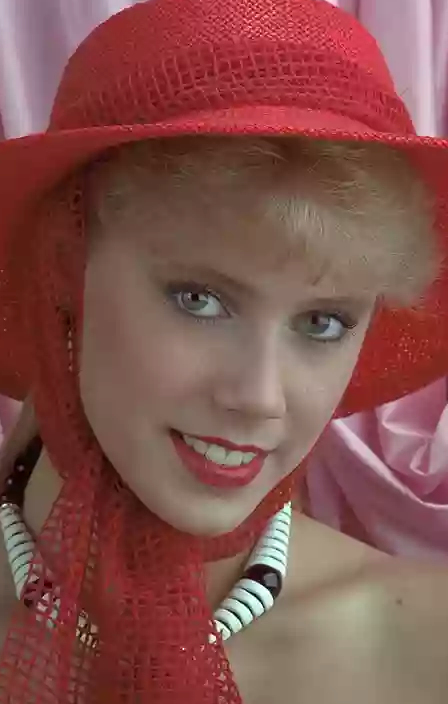}
{\footnotesize\centering 0.1861 BPP\\ \vspace{0.05in}}
\includegraphics[width=0.25\textwidth,trim=0cm 0cm 0cm 0cm,clip]{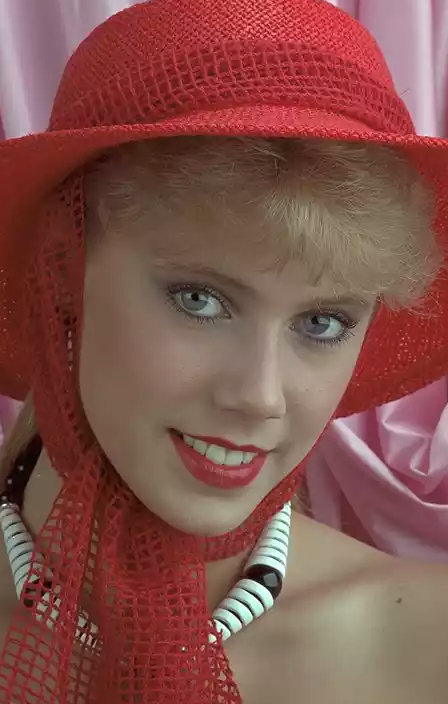}
{\footnotesize\centering 0.4016 BPP\\ \vspace{0.05in}}

{\bf\centering Ours\\ \vspace{0.05in}}
\includegraphics[width=0.25\textwidth,trim=0cm 0cm 0cm 0cm,clip]{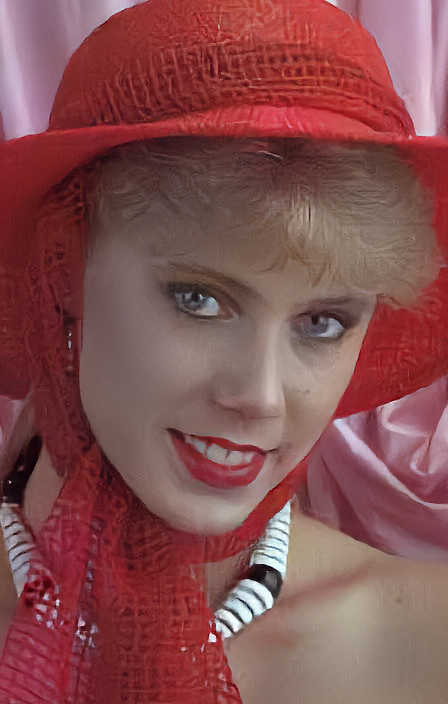}
{\footnotesize\centering 0.0840 BPP\\ \vspace{0.05in}}
\includegraphics[width=0.25\textwidth,trim=0cm 0cm 0cm 0cm,clip]{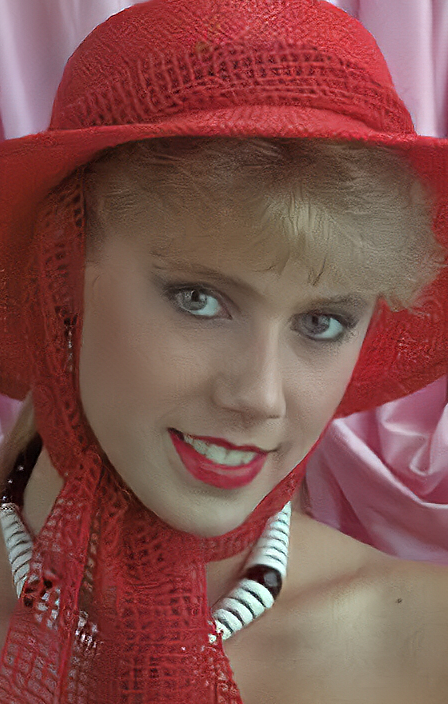}
{\footnotesize\centering 0.1851 BPP\\ \vspace{0.05in}}
\includegraphics[width=0.25\textwidth,trim=0cm 0cm 0cm 0cm,clip]{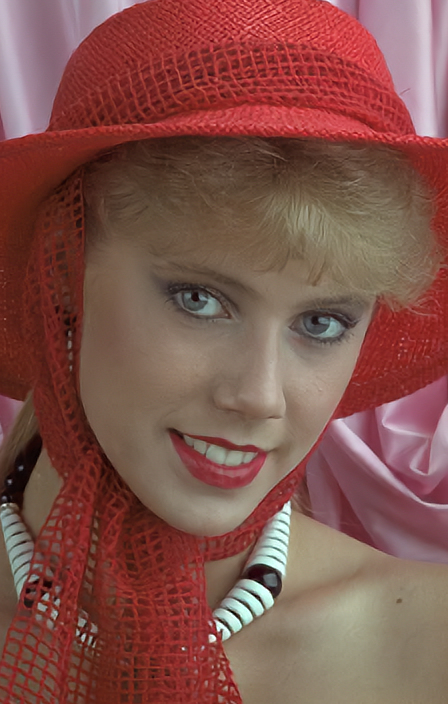}
{\footnotesize\centering 0.3963 BPP\\ \vspace{0.05in}}
\end{multicols}
\end{widepage} 
\end{figure*}

\begin{figure*}[t!]
\vspace{-0.1in}
\begin{widepage}
\begin{multicols}{4}
{\bf\centering JPEG\\ \vspace{0.05in}}
\includegraphics[width=0.25\textwidth,trim=0cm 0cm 0cm 0cm,clip]{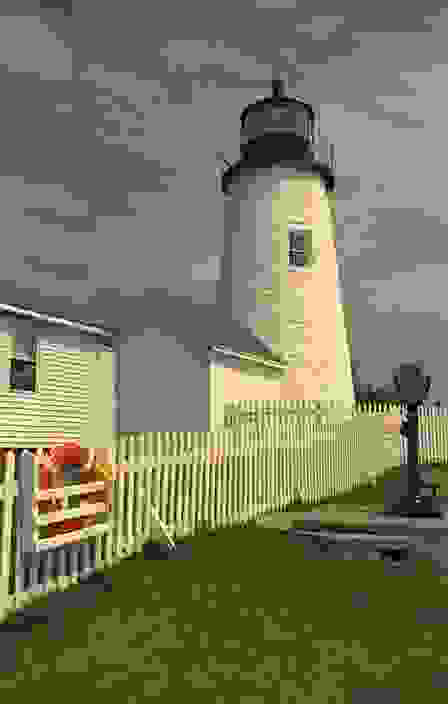}
{\footnotesize\centering 0.0949 BPP\\ \vspace{0.05in}}
\includegraphics[width=0.25\textwidth,trim=0cm 0cm 0cm 0cm,clip]{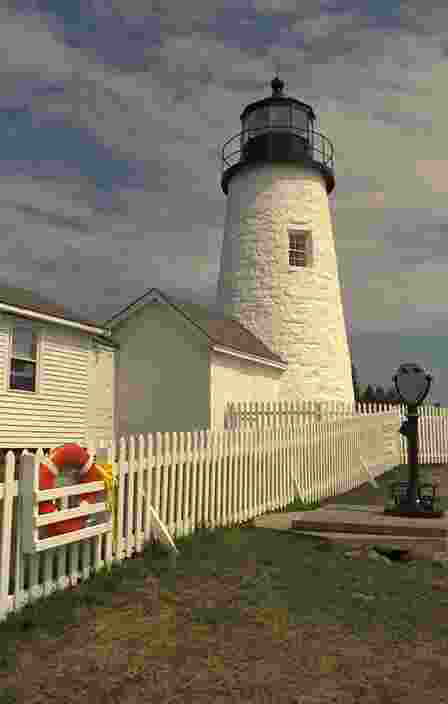}
{\footnotesize\centering 0.1970 BPP\\ \vspace{0.05in}}
\includegraphics[width=0.25\textwidth,trim=0cm 0cm 0cm 0cm,clip]{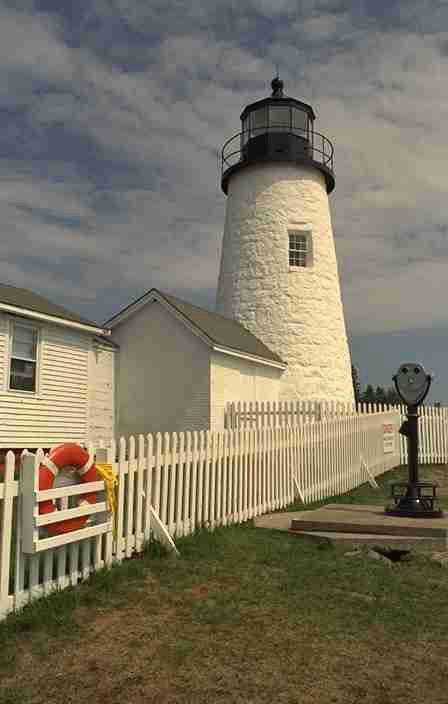}
{\footnotesize\centering 0.4196 BPP\\ \vspace{0.05in}}

{\bf\centering JPEG 2000\\ \vspace{0.05in}}
\includegraphics[width=0.25\textwidth,trim=0cm 0cm 0cm 0cm,clip]{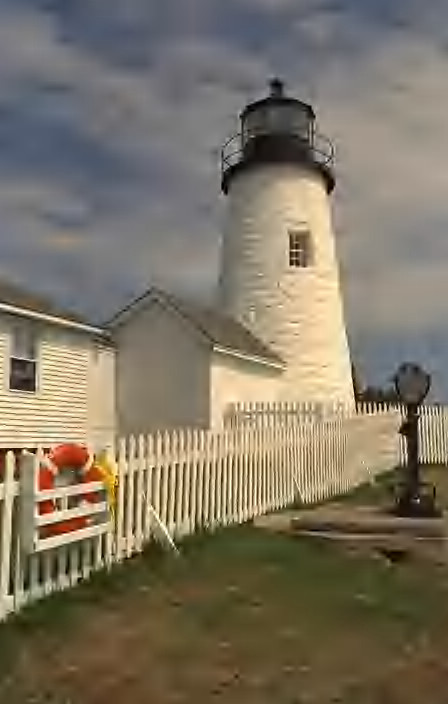}
{\footnotesize\centering 0.0941 BPP\\ \vspace{0.05in}}
\includegraphics[width=0.25\textwidth,trim=0cm 0cm 0cm 0cm,clip]{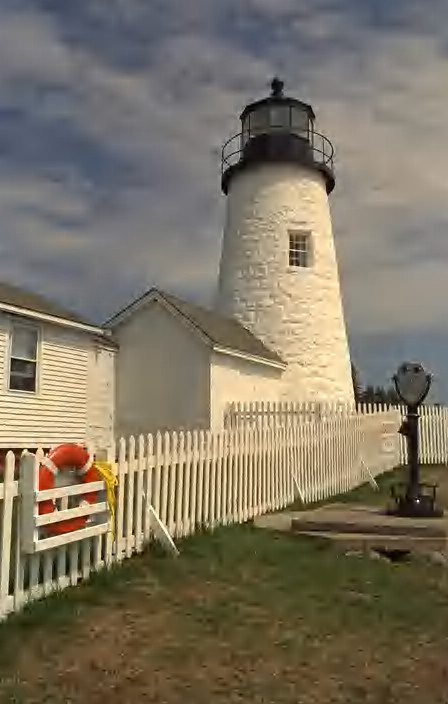}
{\footnotesize\centering 0.1953 BPP\\ \vspace{0.05in}}
\includegraphics[width=0.25\textwidth,trim=0cm 0cm 0cm 0cm,clip]{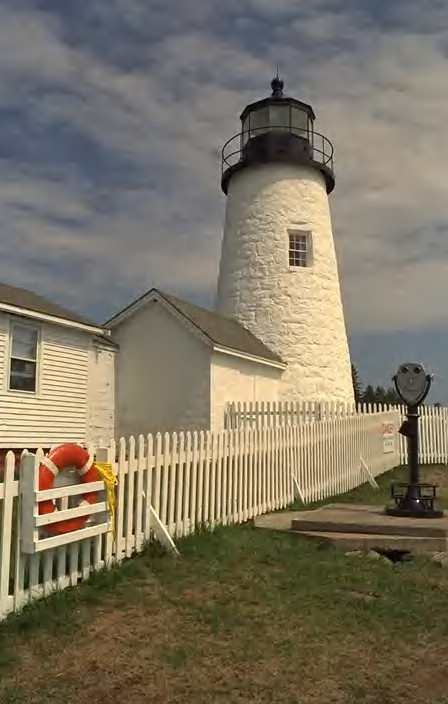}
{\footnotesize\centering 0.4069 BPP\\ \vspace{0.05in}}

{\bf\centering WebP\\ \vspace{0.05in}}
\includegraphics[width=0.25\textwidth,trim=0cm 0cm 0cm 0cm,clip]{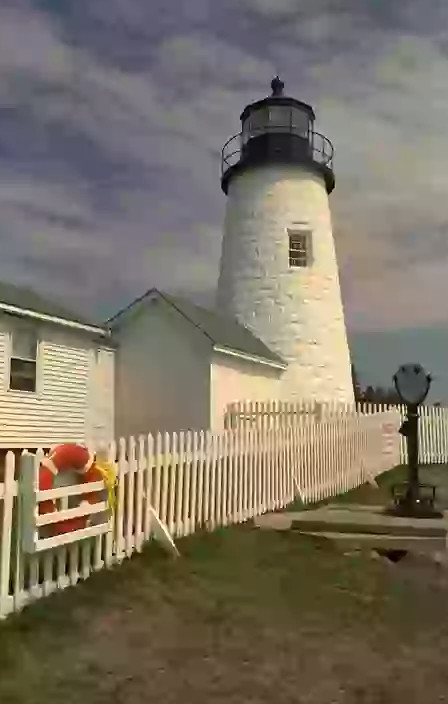}
{\footnotesize\centering 0.1452 BPP\\ \vspace{0.05in}}
\includegraphics[width=0.25\textwidth,trim=0cm 0cm 0cm 0cm,clip]{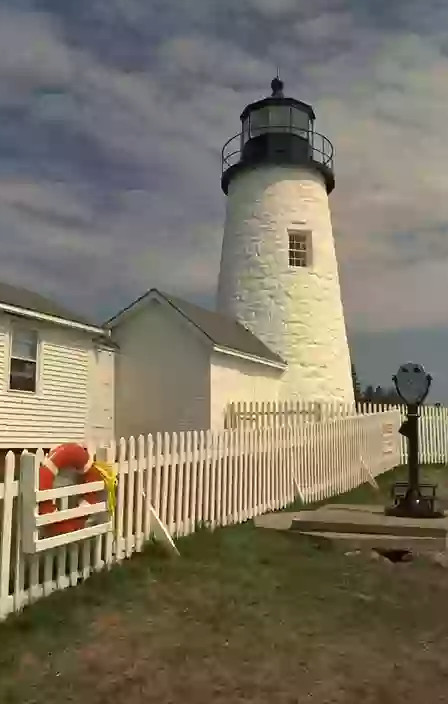}
{\footnotesize\centering 0.1956 BPP\\ \vspace{0.05in}}
\includegraphics[width=0.25\textwidth,trim=0cm 0cm 0cm 0cm,clip]{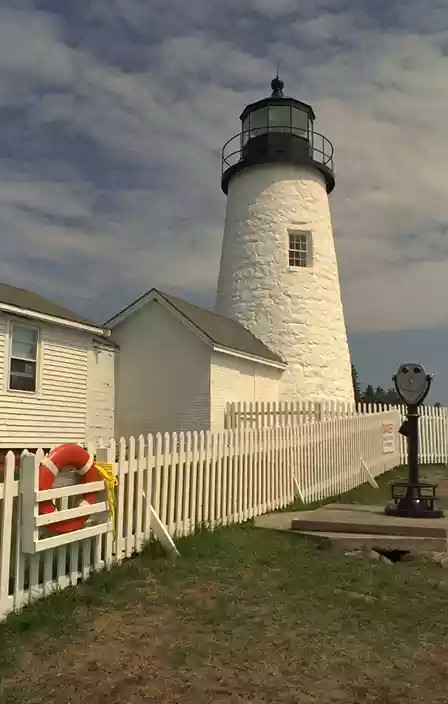}
{\footnotesize\centering 0.4117 BPP\\ \vspace{0.05in}}

{\bf\centering Ours\\ \vspace{0.05in}}
\includegraphics[width=0.25\textwidth,trim=0cm 0cm 0cm 0cm,clip]{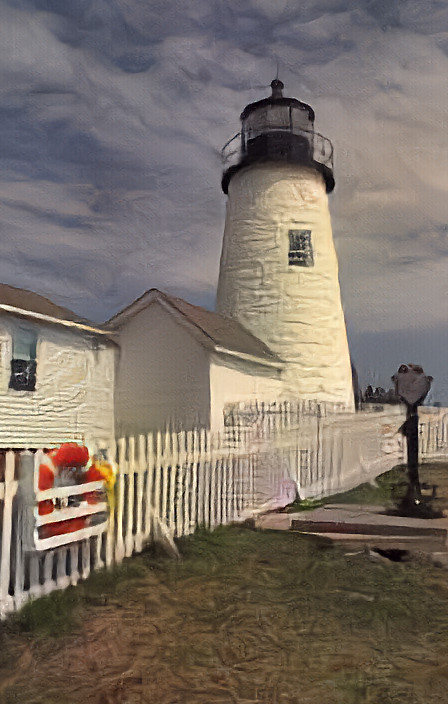}
{\footnotesize\centering 0.0928 BPP\\ \vspace{0.05in}}
\includegraphics[width=0.25\textwidth,trim=0cm 0cm 0cm 0cm,clip]{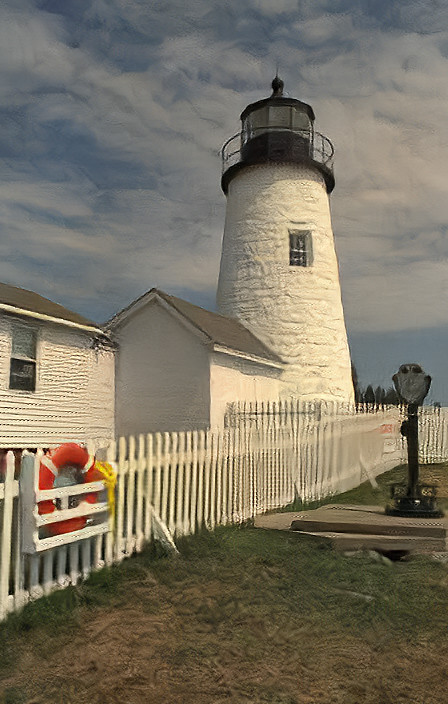}
{\footnotesize\centering 0.1939 BPP\\ \vspace{0.05in}}
\includegraphics[width=0.25\textwidth,trim=0cm 0cm 0cm 0cm,clip]{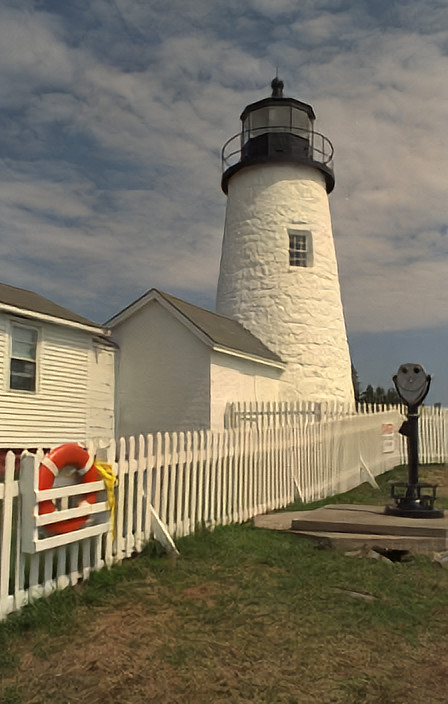}
{\footnotesize\centering 0.4035 BPP\\ \vspace{0.05in}}
\end{multicols}
\end{widepage} 
\end{figure*}

\begin{figure*}[t!]
\vspace{-0.1in}
\begin{widepage}
\begin{multicols}{4}
{\bf\centering JPEG\\ \vspace{0.05in}}
\includegraphics[width=0.25\textwidth,trim=0cm 0cm 0cm 0cm,clip]{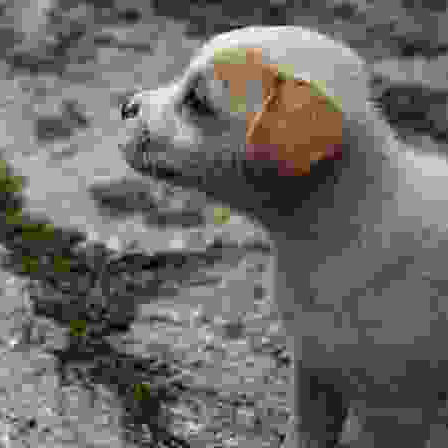}
{\footnotesize\centering 0.1008 BPP\\ \vspace{0.05in}}
\includegraphics[width=0.25\textwidth,trim=0cm 0cm 0cm 0cm,clip]{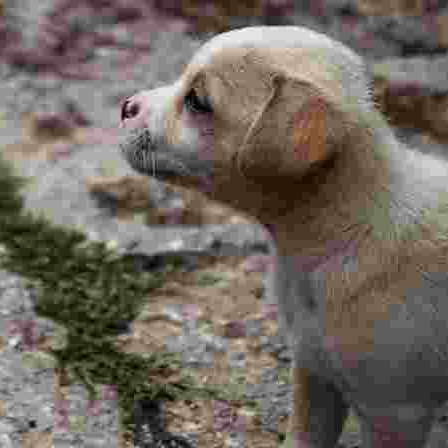}
{\footnotesize\centering 0.2083 BPP\\ \vspace{0.05in}}
\includegraphics[width=0.25\textwidth,trim=0cm 0cm 0cm 0cm,clip]{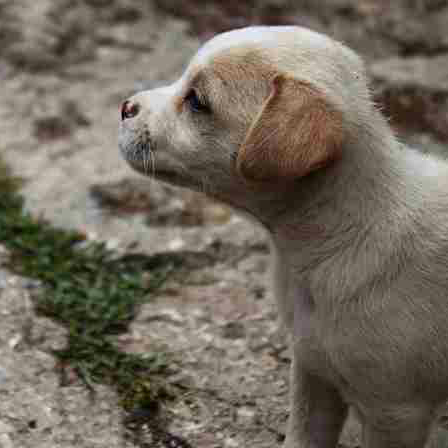}
{\footnotesize\centering 0.3734 BPP\\ \vspace{0.05in}}

{\bf\centering JPEG 2000\\ \vspace{0.05in}}
\includegraphics[width=0.25\textwidth,trim=0cm 0cm 0cm 0cm,clip]{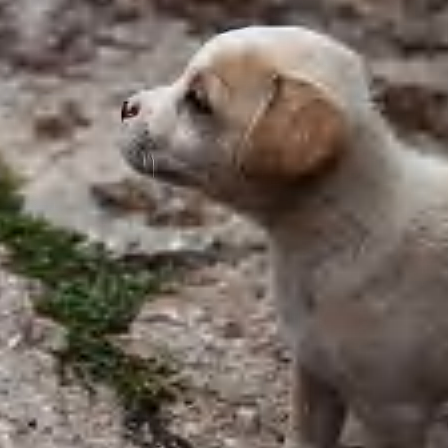}
{\footnotesize\centering 0.0953 BPP\\ \vspace{0.05in}}
\includegraphics[width=0.25\textwidth,trim=0cm 0cm 0cm 0cm,clip]{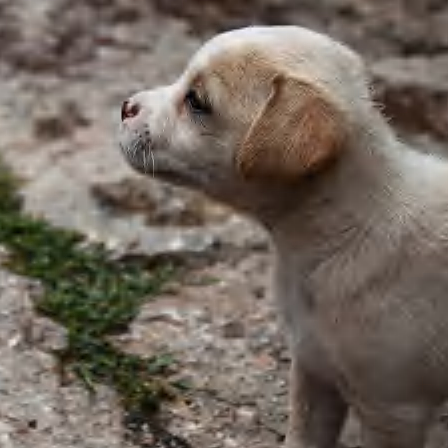}
{\footnotesize\centering 0.1939 BPP\\ \vspace{0.05in}}
\includegraphics[width=0.25\textwidth,trim=0cm 0cm 0cm 0cm,clip]{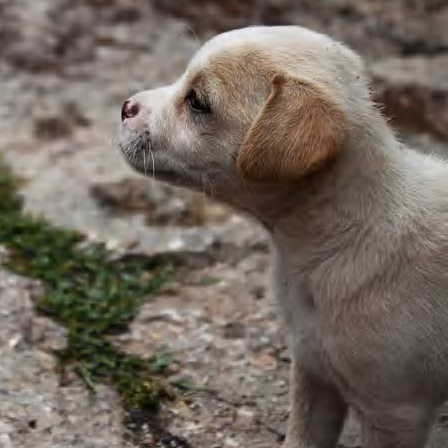}
{\footnotesize\centering 0.3690 BPP\\ \vspace{0.05in}}

{\bf\centering WebP\\ \vspace{0.05in}}
\includegraphics[width=0.25\textwidth,trim=0cm 0cm 0cm 0cm,clip]{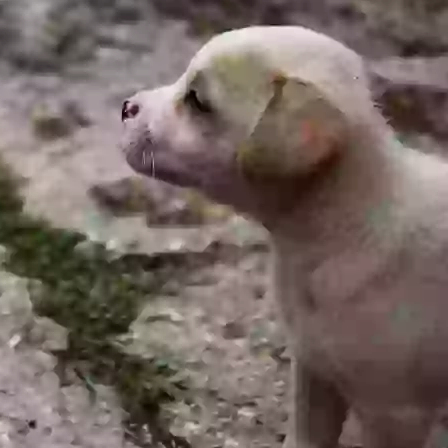}
{\footnotesize\centering 0.1392 BPP\\ \vspace{0.05in}}
\includegraphics[width=0.25\textwidth,trim=0cm 0cm 0cm 0cm,clip]{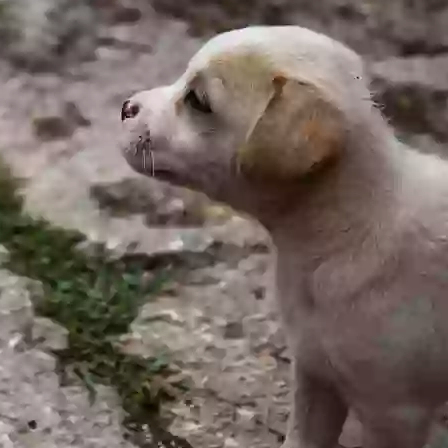}
{\footnotesize\centering 0.1973 BPP\\ \vspace{0.05in}}
\includegraphics[width=0.25\textwidth,trim=0cm 0cm 0cm 0cm,clip]{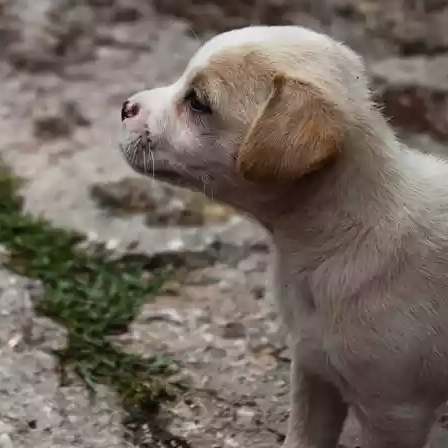}
{\footnotesize\centering 0.3672 BPP\\ \vspace{0.05in}}

{\bf\centering Ours\\ \vspace{0.05in}}
\includegraphics[width=0.25\textwidth,trim=0cm 0cm 0cm 0cm,clip]{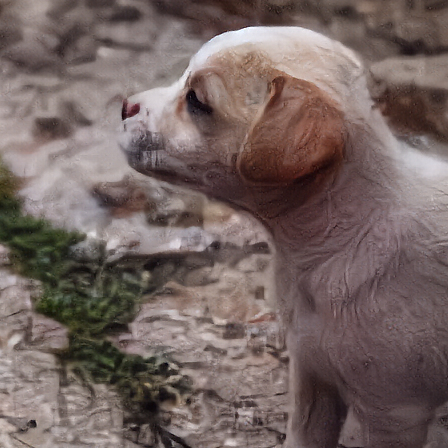}
{\footnotesize\centering 0.0949 BPP\\ \vspace{0.05in}}
\includegraphics[width=0.25\textwidth,trim=0cm 0cm 0cm 0cm,clip]{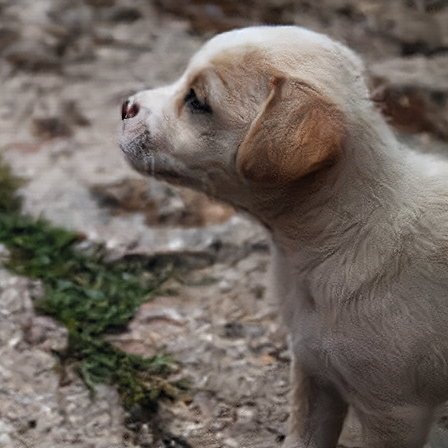}
{\footnotesize\centering 0.1921 BPP\\ \vspace{0.05in}}
\includegraphics[width=0.25\textwidth,trim=0cm 0cm 0cm 0cm,clip]{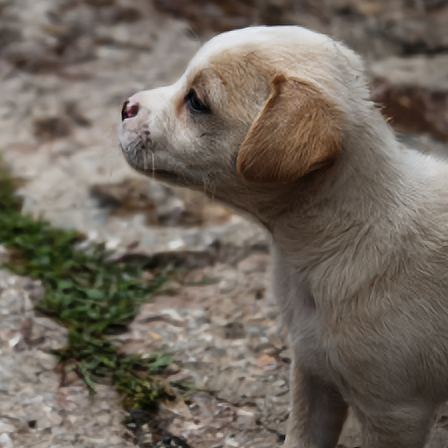}
{\footnotesize\centering 0.3643 BPP\\ \vspace{0.05in}}
\end{multicols}
\end{widepage} 
\end{figure*}

\begin{figure*}[t!]
\vspace{-0.1in}
\begin{widepage}
\begin{multicols}{4}
{\bf\centering JPEG\\ \vspace{0.05in}}
\includegraphics[width=0.25\textwidth,trim=0cm 0cm 0cm 0cm,clip]{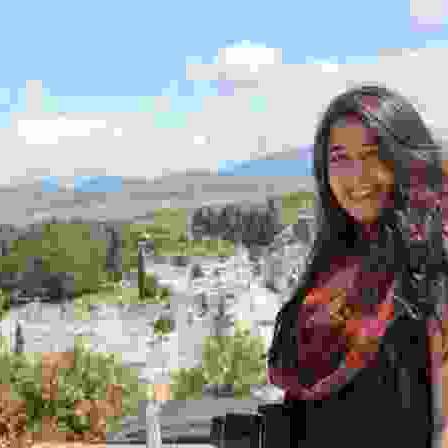}
{\footnotesize\centering 0.1101 BPP\\ \vspace{0.05in}}
\includegraphics[width=0.25\textwidth,trim=0cm 0cm 0cm 0cm,clip]{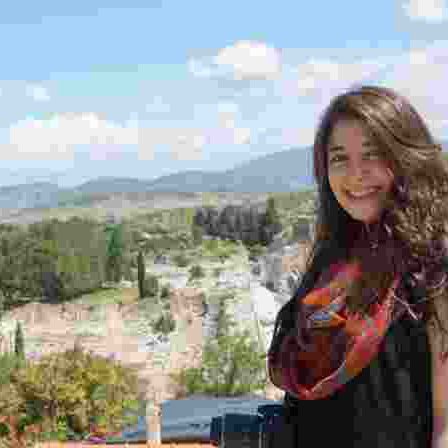}
{\footnotesize\centering 0.2071 BPP\\ \vspace{0.05in}}
\includegraphics[width=0.25\textwidth,trim=0cm 0cm 0cm 0cm,clip]{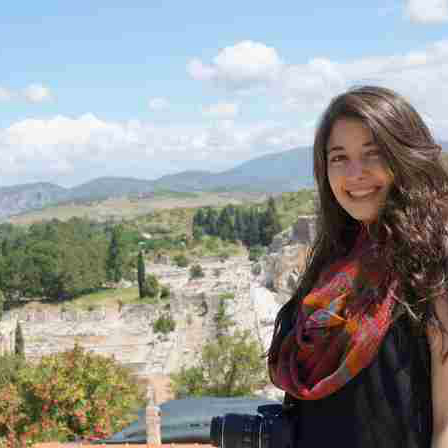}
{\footnotesize\centering 0.4055 BPP\\ \vspace{0.05in}}

{\bf\centering JPEG 2000\\ \vspace{0.05in}}
\includegraphics[width=0.25\textwidth,trim=0cm 0cm 0cm 0cm,clip]{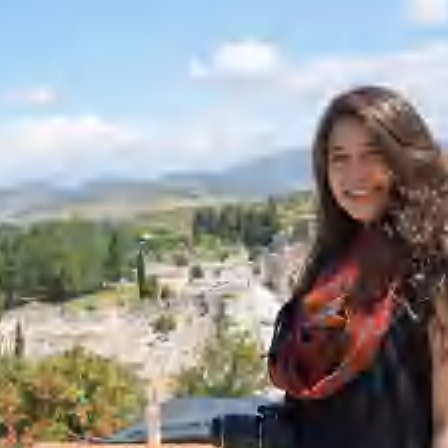}
{\footnotesize\centering 0.0947 BPP\\ \vspace{0.05in}}
\includegraphics[width=0.25\textwidth,trim=0cm 0cm 0cm 0cm,clip]{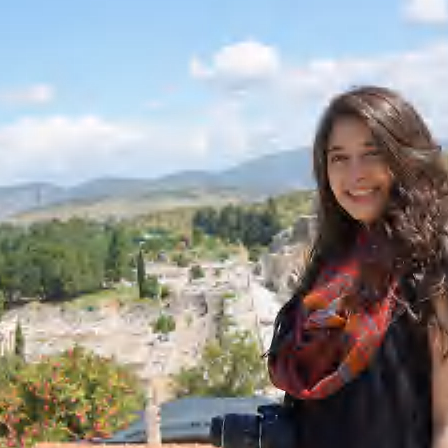}
{\footnotesize\centering 0.2014 BPP\\ \vspace{0.05in}}
\includegraphics[width=0.25\textwidth,trim=0cm 0cm 0cm 0cm,clip]{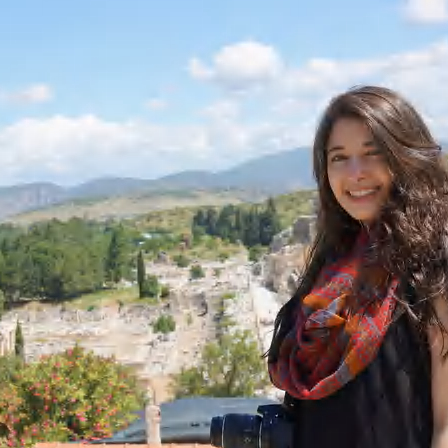}
{\footnotesize\centering 0.4002 BPP\\ \vspace{0.05in}}

{\bf\centering WebP\\ \vspace{0.05in}}
\includegraphics[width=0.25\textwidth,trim=0cm 0cm 0cm 0cm,clip]{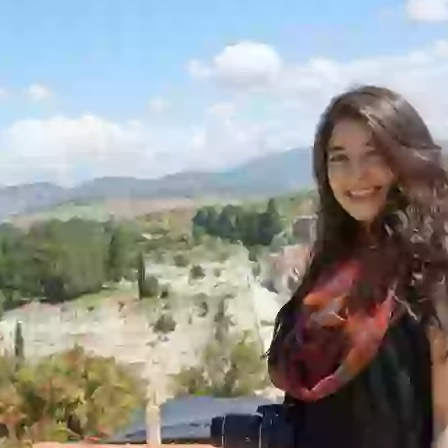}
{\footnotesize\centering 0.1510 BPP\\ \vspace{0.05in}}
\includegraphics[width=0.25\textwidth,trim=0cm 0cm 0cm 0cm,clip]{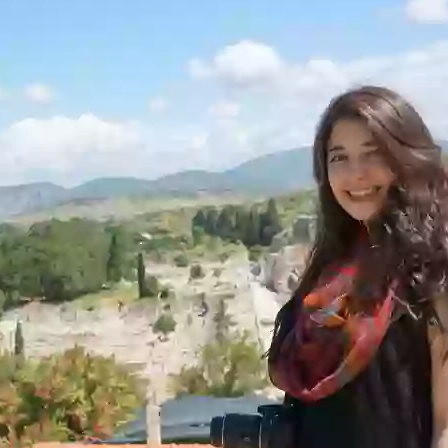}
{\footnotesize\centering 0.1989 BPP\\ \vspace{0.05in}}
\includegraphics[width=0.25\textwidth,trim=0cm 0cm 0cm 0cm,clip]{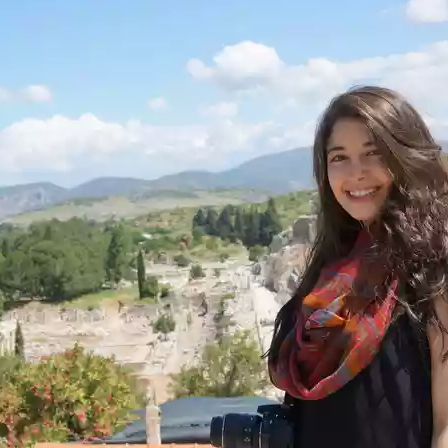}
{\footnotesize\centering 0.4087 BPP\\ \vspace{0.05in}}

{\bf\centering Ours\\ \vspace{0.05in}}
\includegraphics[width=0.25\textwidth,trim=0cm 0cm 0cm 0cm,clip]{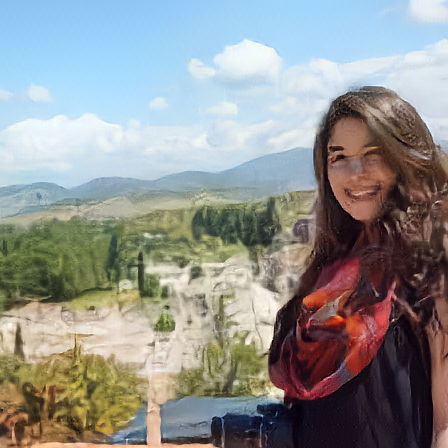}
{\footnotesize\centering 0.0941 BPP\\ \vspace{0.05in}}
\includegraphics[width=0.25\textwidth,trim=0cm 0cm 0cm 0cm,clip]{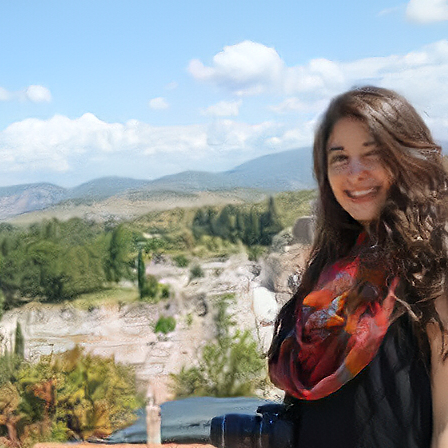}
{\footnotesize\centering 0.1940 BPP\\ \vspace{0.05in}}
\includegraphics[width=0.25\textwidth,trim=0cm 0cm 0cm 0cm,clip]{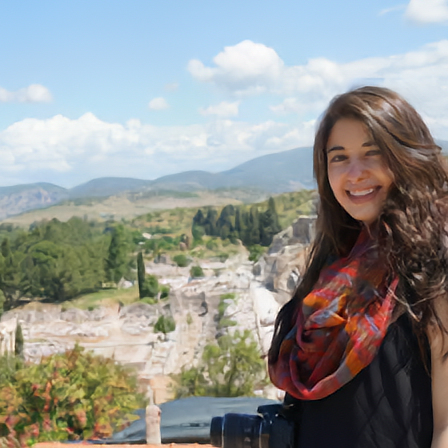}
{\footnotesize\centering 0.3971 BPP\\ \vspace{0.05in}}
\end{multicols}
\end{widepage} 
\end{figure*}

\begin{figure*}[t!]
\vspace{-0.1in}
\begin{widepage}
\begin{multicols}{4}
{\bf\centering JPEG\\ \vspace{0.05in}}
\includegraphics[width=0.25\textwidth,trim=0cm 0cm 0cm 0cm,clip]{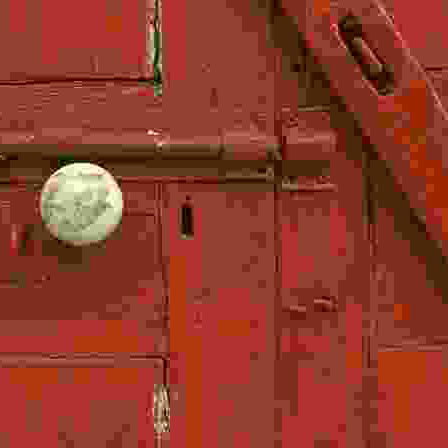}
{\footnotesize\centering 0.0881 BPP\\ \vspace{0.05in}}
\includegraphics[width=0.25\textwidth,trim=0cm 0cm 0cm 0cm,clip]{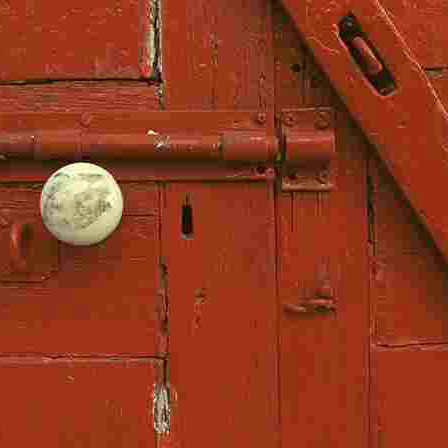}
{\footnotesize\centering 0.1923 BPP\\ \vspace{0.05in}}
\includegraphics[width=0.25\textwidth,trim=0cm 0cm 0cm 0cm,clip]{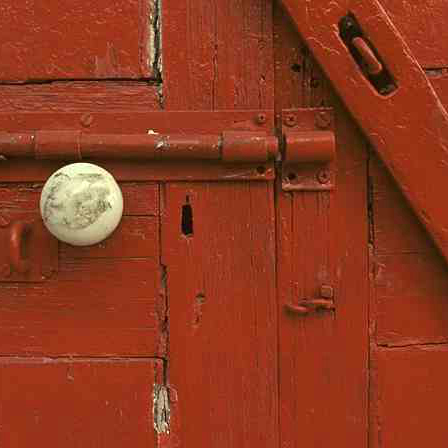}
{\footnotesize\centering 0.4012 BPP\\ \vspace{0.05in}}

{\bf\centering JPEG 2000\\ \vspace{0.05in}}
\includegraphics[width=0.25\textwidth,trim=0cm 0cm 0cm 0cm,clip]{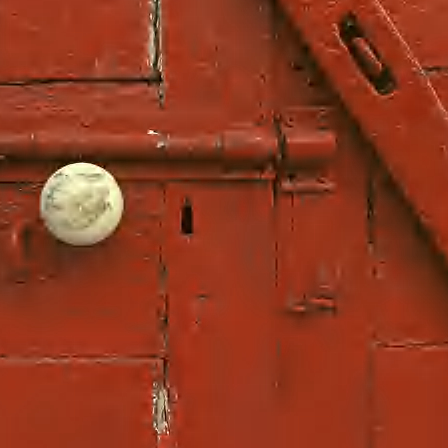}
{\footnotesize\centering 0.0846 BPP\\ \vspace{0.05in}}
\includegraphics[width=0.25\textwidth,trim=0cm 0cm 0cm 0cm,clip]{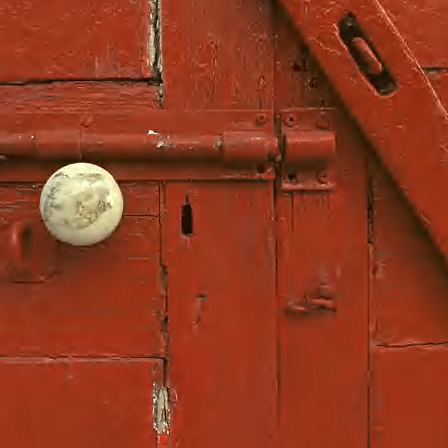}
{\footnotesize\centering 0.1889 BPP\\ \vspace{0.05in}}
\includegraphics[width=0.25\textwidth,trim=0cm 0cm 0cm 0cm,clip]{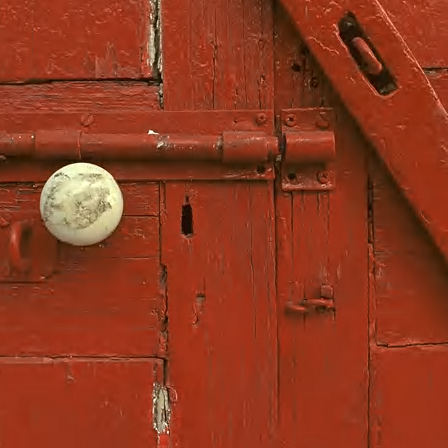}
{\footnotesize\centering 0.4002 BPP\\ \vspace{0.05in}}

{\bf\centering WebP\\ \vspace{0.05in}}
\includegraphics[width=0.25\textwidth,trim=0cm 0cm 0cm 0cm,clip]{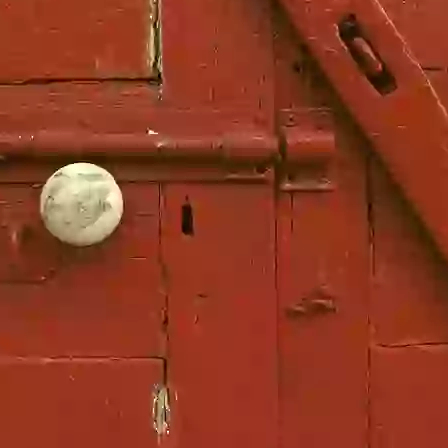}
{\footnotesize\centering 0.0841 BPP\\ \vspace{0.05in}}
\includegraphics[width=0.25\textwidth,trim=0cm 0cm 0cm 0cm,clip]{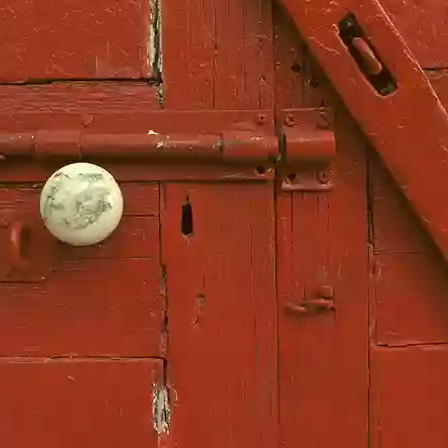}
{\footnotesize\centering 0.1952 BPP\\ \vspace{0.05in}}
\includegraphics[width=0.25\textwidth,trim=0cm 0cm 0cm 0cm,clip]{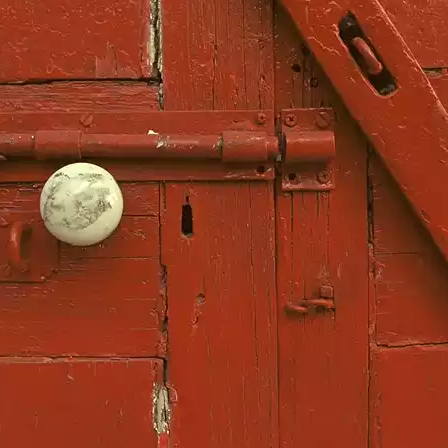}
{\footnotesize\centering 0.4047 BPP\\ \vspace{0.05in}}

{\bf\centering Ours\\ \vspace{0.05in}}
\includegraphics[width=0.25\textwidth,trim=0cm 0cm 0cm 0cm,clip]{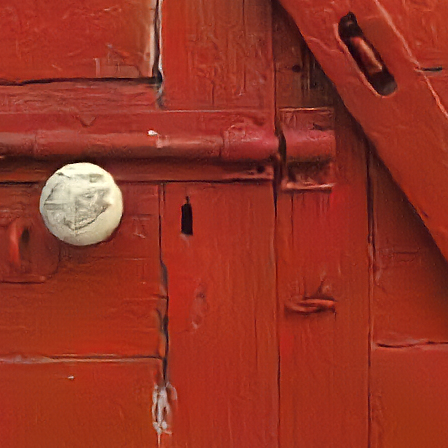}
{\footnotesize\centering 0.0828 BPP\\ \vspace{0.05in}}
\includegraphics[width=0.25\textwidth,trim=0cm 0cm 0cm 0cm,clip]{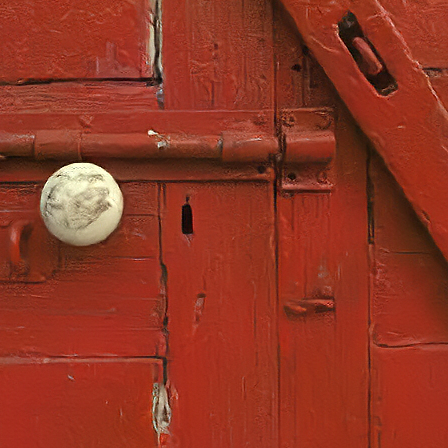}
{\footnotesize\centering 0.1885 BPP\\ \vspace{0.05in}}
\includegraphics[width=0.25\textwidth,trim=0cm 0cm 0cm 0cm,clip]{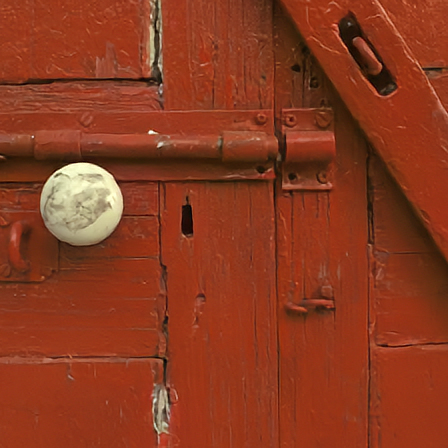}
{\footnotesize\centering 0.3996 BPP\\ \vspace{0.05in}}
\end{multicols}
\end{widepage} 
\end{figure*}
\end{appendices}

\end{document}